%% file: pddm.tex
\documentclass{article}
\usepackage{times}
\usepackage{graphicx}
\usepackage{wrapfig}
\usepackage{amsmath}
\usepackage{gensymb} 
\usepackage[linewidth=0.5pt,linecolor=black]{mdframed} 
\usepackage{algorithm}
\usepackage{algorithmic}
\usepackage{bbm}
\usepackage{multirow}
\usepackage[bottom]{footmisc}

\newlength{\bibitemsep}
\newlength{\bibparskip}
\let\oldthebibliography\thebibliography
\renewcommand\thebibliography[1]{%
\oldthebibliography{#1}%
\setlength{\parskip}{\bibitemsep}%
\setlength{\itemsep}{\bibparskip}%
}
\usepackage[compact]{titlesec} 
\titlespacing{\section}{0pt}{*0}{*0}
\usepackage[final]{pddm_2019}

\title{Deep Dynamics Models \\for Learning Dexterous Manipulation}
\date{\vspace{-5ex}}

\author{
    Anusha Nagabandi, Kurt Konoglie, Sergey Levine, Vikash Kumar \\
    Google Brain\\
}

\newcommand{\ourmethod}{PDDM}

\begin{document}
\maketitle

\begin{abstract}
Dexterous multi-fingered hands can provide robots with the ability to flexibly perform a wide range of manipulation skills. However, many of the more complex behaviors are also notoriously difficult to control: Performing in-hand object manipulation, executing finger gaits to move objects, and exhibiting precise fine motor skills such as writing, all require finely balancing contact forces, breaking and reestablishing contacts repeatedly, and maintaining control of unactuated objects. 
Learning-based techniques provide the appealing possibility of acquiring these skills directly from data, but current learning approaches either require large amounts of data and produce task-specific policies, or they have not yet been shown to scale up to more complex and realistic tasks requiring fine motor skills. In this work, we demonstrate that our method of online planning with deep dynamics models (\ourmethod) addresses both of these limitations; we show that improvements in learned dynamics models, together with improvements in online model-predictive control, can indeed enable efficient and effective learning of flexible contact-rich dexterous manipulation skills -- and that too, on a 24-DoF anthropomorphic hand in the real world, using just 4 hours of purely real-world data to learn to simultaneously coordinate multiple free-floating objects. Videos can be found at \url{https://sites.google.com/view/pddm/}
\end{abstract}

\vspace{-3mm}
\keywords{Manipulation, Model-based learning, Robots}

\input{01_introduction.tex}
\input{03_related.tex}
\input{04_method.tex}
\input{05_results.tex}
\input{06_conclusion.tex}

\clearpage
\bibliography{pddm.bib}

\newpage
\appendix
\input{appendix}

\end{document}

%% file: 01_introduction.tex
\vspace{-2mm}
\section{Introduction}
\label{sec:intro}
\vspace{-2mm}

Dexterous manipulation with multi-fingered hands represents a grand challenge in robotics: the versatility of the human hand is as yet unrivaled by the capabilities of robotic systems, and bridging this gap will enable more general and capable robots. Although some real-world tasks can be accomplished with simple parallel jaw grippers, there are countless tasks in which dexterity in the form of redundant degrees of freedom is critical. In fact, dexterous manipulation is defined~\cite{okamura2000overview} as being object-centric, with the goal of controlling object movement through precise control of forces and motions -- something that is not possible without the ability to simultaneously impact the object from multiple directions. Through added controllability and stability, multi-fingered hands enable useful fine motor skills that are necessary for deliberate interaction with objects. For example, using only two fingers to attempt common tasks such as opening the lid of a jar, hitting a nail with a hammer, or writing on paper with a pencil would quickly encounter the challenges of slippage, complex contact forces, and underactuation. Success in such settings requires a sufficiently dexterous hand, as well as an intelligent policy that can endow such a hand with the appropriate control strategy.

\begin{figure}
\centering
\includegraphics[width=0.9\columnwidth]{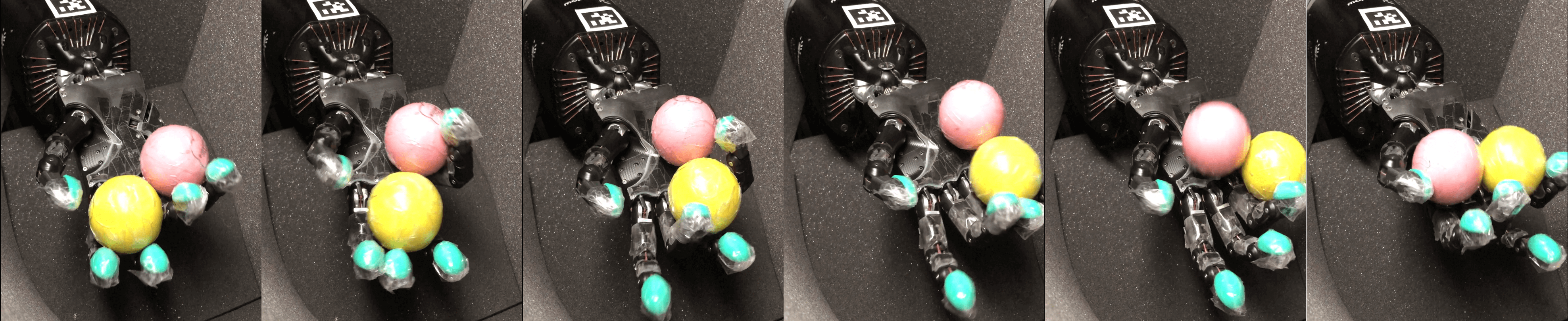}
\vspace{-2mm}
\caption{\footnotesize{\ourmethod ~can efficiently and effectively learn complex dexterous manipulation skills in both simulation and the real world. Here, the learned model uses less than 4 hours of experience to enable the Shadow Hand to rotate two free-floating Baoding balls in the palm without any prior knowledge of system dynamics.}}
\label{fig:teaser_baoding}
\end{figure}

The principle challenges in dexterous manipulation stem from the need to coordinate numerous joints and impart complex forces onto the object of interest. The need to repeatedly establish and break contacts presents an especially difficult problem for analytic approaches, which require accurate models of the physics of the system.
Learning offers a promising data-driven alternative. Model-free reinforcement learning (RL) methods can learn policies that achieve good performance on complex tasks~\citep{van2015learning,levine2016end,rajeswaran2017learning}; however, we will show that these state-of-the-art algorithms struggle when a high degree of flexibility is required, such as moving a pencil to follow \emph{arbitrary} user-specified strokes. Here, complex contact dynamics and high chances of task failure make the overall skill much more difficult. Model-free methods also require large amounts of data, making them difficult to use in the real world. Model-based RL methods, on the other hand, can be much more efficient, but have not yet been scaled up to such complex tasks. In this work, we aim to push the boundary on this task complexity; consider, for instance, the task of rotating two Baoding balls around the palm of your hand (\autoref{fig:teaser_baoding}). We will discuss how model-based RL methods can solve such tasks, both in simulation and on a real-world robot.

\begin{wrapfigure}{r}{0.42\columnwidth}
\vspace{-0.17in}
\centering
\includegraphics[width=0.18\columnwidth]{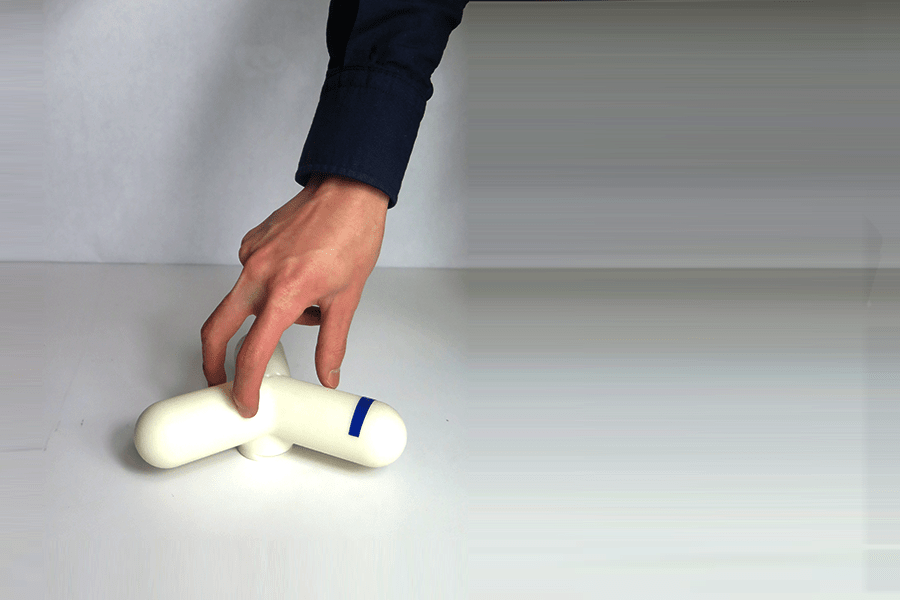}
\includegraphics[width=0.18\columnwidth,trim={10cm 8.5cm 16cm 8.5cm},clip]{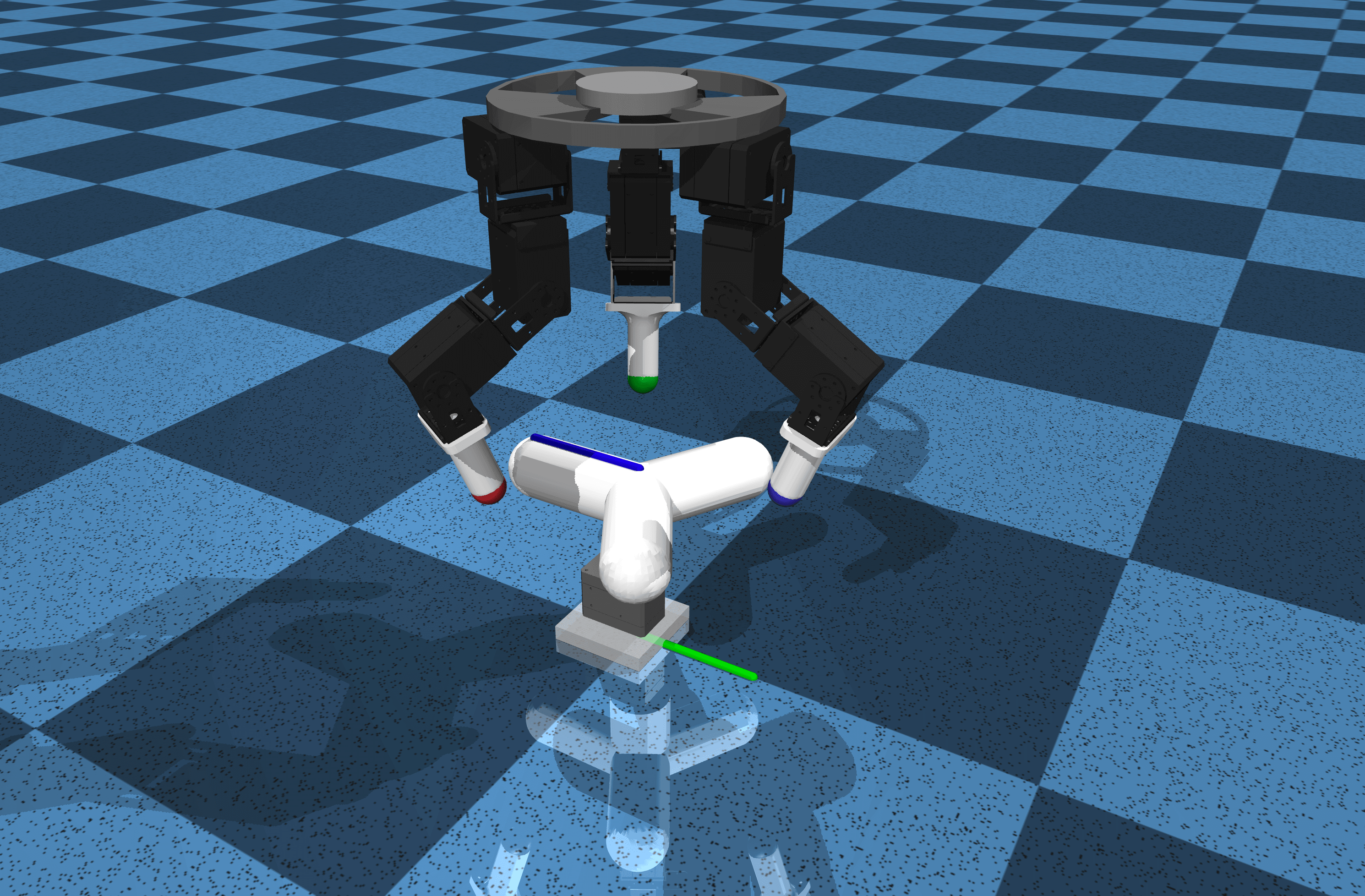}\\
\includegraphics[width=0.18\columnwidth]{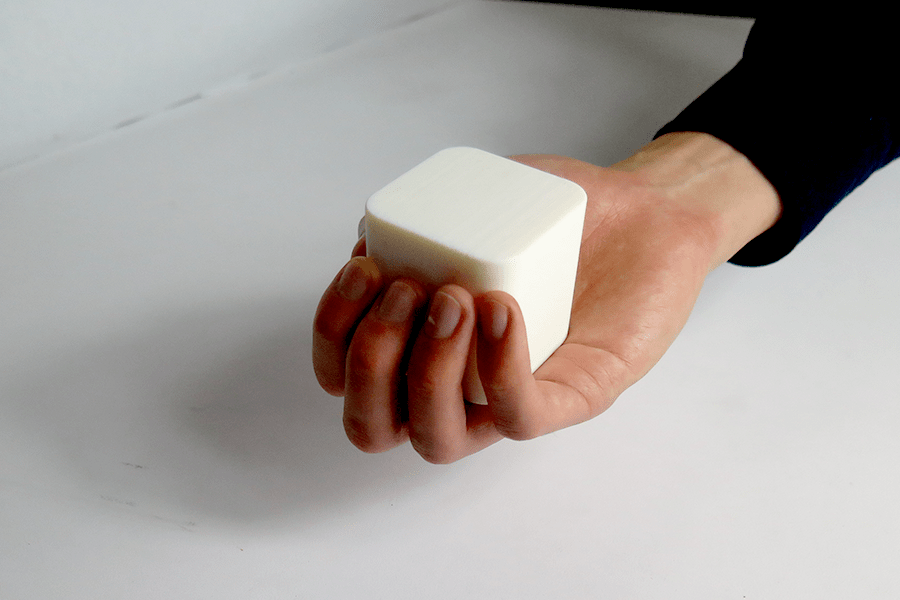}
\includegraphics[width=0.18\columnwidth]{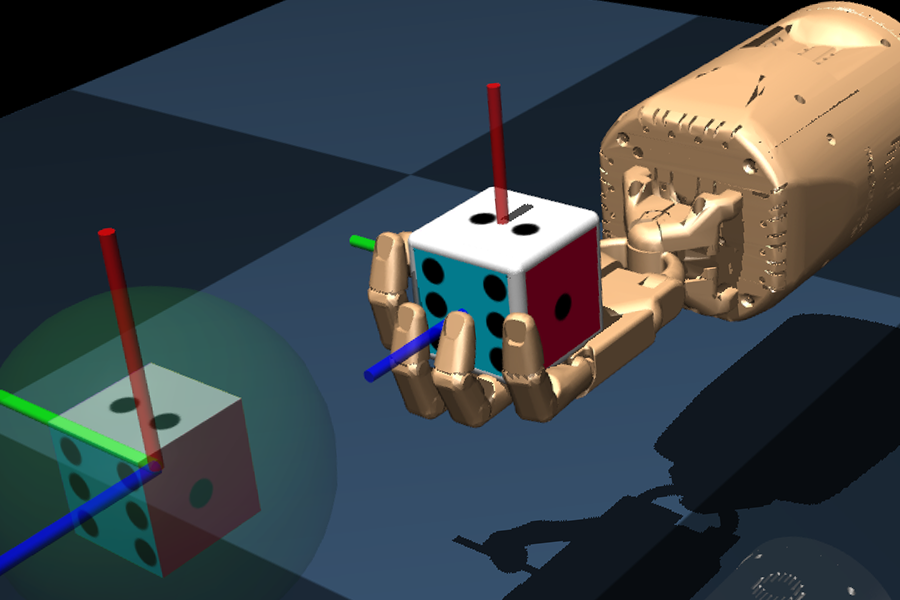}\\
\includegraphics[width=0.18\columnwidth]{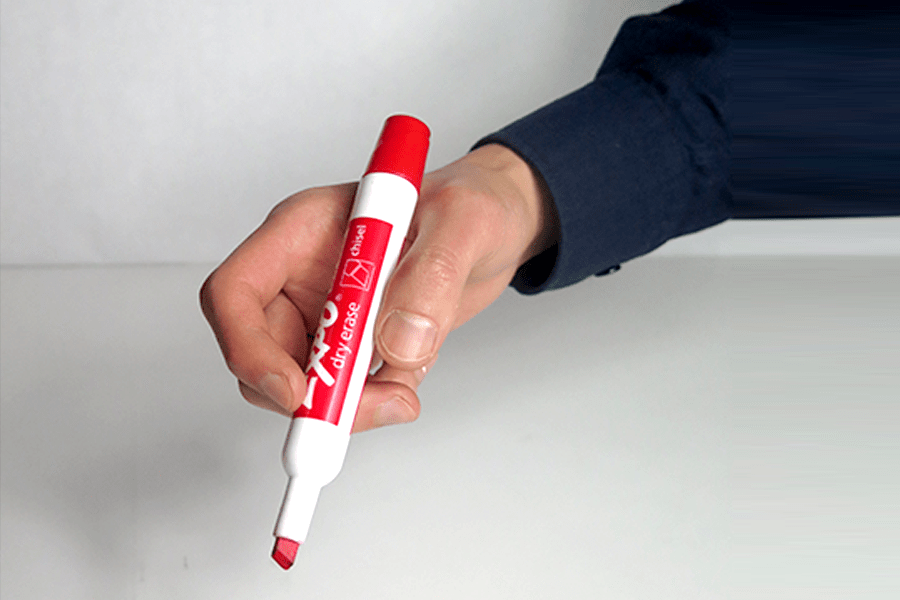}
\includegraphics[width=0.18\columnwidth]{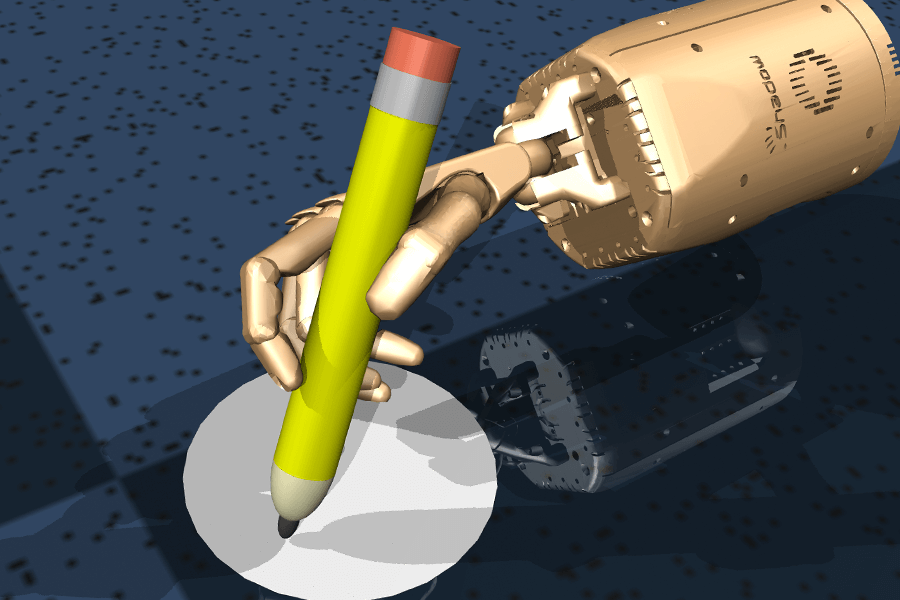}\\
\includegraphics[width=0.18\columnwidth]{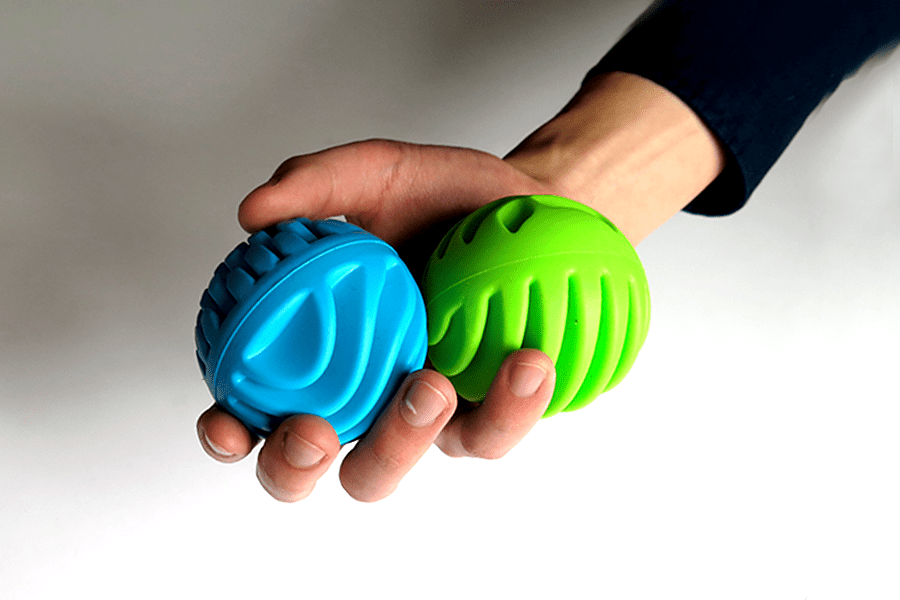}
\includegraphics[width=0.18\columnwidth]{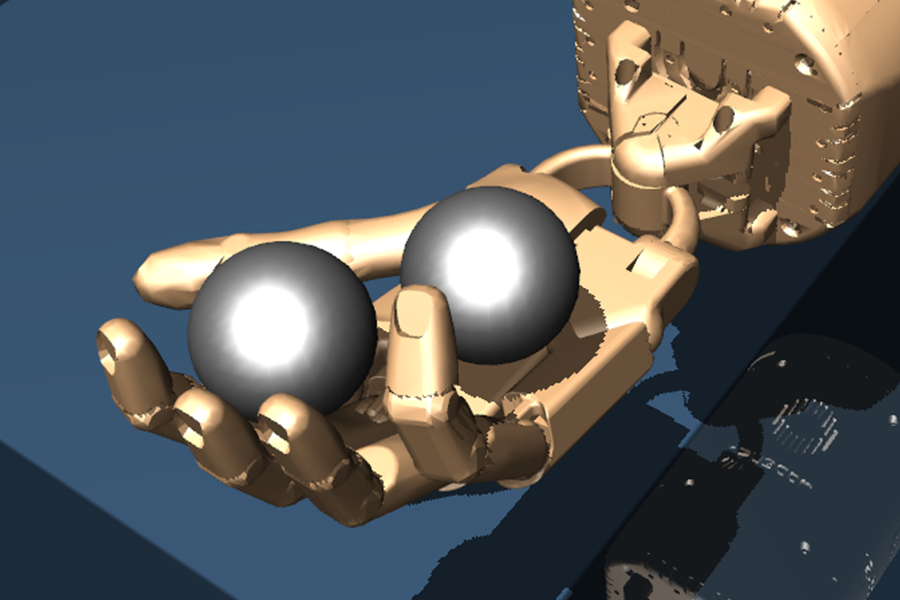}\\
\includegraphics[width=0.18\columnwidth]{figs/task_human/task_baoding2.png}
\scalebox{-1}[1]{\includegraphics[width=0.18\columnwidth,trim={1.5cm 0 0 0},clip]{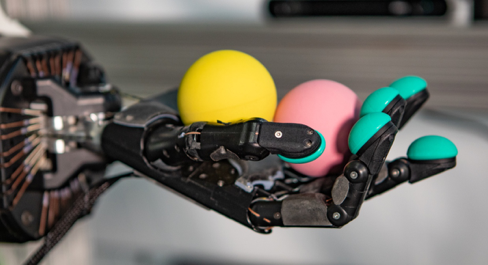}}
\vspace{-2mm}
\caption{Task suite of simulated and real-world dexterous manipulation: valve rotation, in-hand reorientation, handwriting, and manipulating Baoding balls.}
\vspace{-7mm}
\label{fig:tasks}
\end{wrapfigure}

Algorithmically, we present a technique that combines elements of recently-developed uncertainty-aware neural network models with state-of-the-art gradient-free trajectory optimization. While the individual components of our method are based heavily on prior work, we show that their combination is both novel and critical. Our approach, based on deep model-based RL, challenges the general machine learning community's notion that models are difficult to learn and do not yet deliver control results that are as impressive as model-free methods. In this work, we push forward the empirical results of model-based RL, in both simulation and the real world, on a suite of dexterous manipulation tasks starting with a 9-DoF three-fingered hand \cite{zhu2019dexterous} rotating a valve, and scaling up to a 24-DoF anthropomorphic hand executing handwriting and manipulating free-floating objects (\autoref{fig:tasks}). These realistic tasks require not only learning about interactions between the robot and objects in the world, but also effective planning to find precise and coordinated maneuvers while avoiding task failure (e.g., dropping objects). To the best of our knowledge, our paper demonstrates for the first time that deep neural network models can indeed enable sample-efficient and autonomous discovery of fine motor skills with high-dimensional manipulators, including a real-world dexterous hand trained entirely using just 4 hours of real-world data.

%% file: 03_related.tex
\vspace{-3mm}
\section{Related Work}
\label{sec:related}
\vspace{-3mm}

In recent years, the mechanical and hardware development of multi-fingered robotic hands has significantly advanced ~\citep{butterfass2001dlr, xu2016design}, but our manipulation capabilities haven't scaled similarly. Prior work in this area has explored a wide spectrum of manipulation strategies: \citep{sundaralingam2018geometric}'s optimizer used geometric meshes of the object to plan finger-gaiting maneuvers, \citep{andrews2013goal}'s multi-phase planner used the help of offline simulations and a reduced basis of hand poses to accelerate the parameter search for in-hand rotation of a sphere, \citep{dogar2010push}'s motion planning algorithm used the mechanics of pushing to funnel an object into a stable grasp state, and \citep{bai2014dexterous}'s controller used the conservation of mechanical energy to control the tilt of a palm and roll objects to desired positions on the hand. These types of manipulation techniques have thus far struggled to scale to more complex tasks or sophisticated manipulators in simulation as well as the real world, perhaps due to their need for precise characterization of the system and its environment. Reasoning through contact models and motions cones~\citep{chavan2018hand, kolbert2016experimental}, for example, requires computation time that scales exponentially with the number of contacts, and has thus been limited to simpler manipulators and more controlled tasks. In this work, we aim to significantly scale up the complexity of feasible tasks, while also minimizing such task-specific formulations.

More recent work in deep RL has studied this question through the use of data-driven learning to make sense of observed phenomenon~\citep{andrychowicz2018learning, van2015learning}. These methods, while powerful, require large amounts of system interaction to learn successful control policies, making them difficult to apply in the real world. Some work~\citep{rajeswaran2017learning,zhu2019dexterous} has used expert demonstrations to improve this sample efficiency. In contrast, our method is sample efficient without requiring any expert demonstrations, and is still able to leverage data-driven learning techniques to acquire challenging dexterous manipulation skills.

Model-based RL has the potential to provide both efficient and flexible learning. In fact, methods that assume perfect knowledge of system dynamics can achieve very impressive manipulation behaviors~\citep{mordatch2012contact, lowrey2018plan} using generally applicable learning and control techniques. Other work has focused on learning these models using high-capacity function approximators~\citep{deisenroth2013survey,lenz2015deepmpc,levine2016end,nagabandi2017roach,nagabandi2018mbrl,williams2017information,chua2018pets} and probabilistic dynamics models~\citep{Deisenroth2011_ICML,ko2008gp, deisenroth2012toward, doerr2017mbpid}. Our method combines components from multiple prior works, including uncertainty estimation~\citep{Deisenroth2011_ICML,chua2018pets,kurutach2018model}, deep models and model-predictive control (MPC) ~\citep{nagabandi2017roach}, and stochastic optimization for planning~\citep{williams15mppi}. Model-based RL methods, including recent work in uncertainty estimation~\citep{malik2019calibrated} and combining policy networks with online planning~\citep{wang2019exploring}, have unfortunately mostly been studied and shown on lower-dimensional (and often, simulated) benchmark tasks, and scaling these methods to higher dimensional tasks such as dexterous manipulation has proven to be a challenge. As illustrated in our evaluation, the particular synthesis of different ideas in this work allows model-based RL to push forward the task complexity of achievable dexterous manipulation skills, and to extend this progress to even a real-world robotic hand.

%% file: 04_method.tex
\vspace{-2mm}
\section{Deep Models for Dexterous Manipulation}
\vspace{-2mm}

In order to leverage the benefits of autonomous learning from data-driven methods while also enabling efficient and flexible task execution, we extend deep model-based RL approaches to the domain of dexterous manipulation. Our method of online planning with deep dynamics models (\ourmethod) builds on prior work that uses MPC with deep models~\cite{nagabandi2018mbrl} and ensembles for model uncertainty estimation~\cite{chua2018pets}. However, as we illustrate in our experiments, the particular combination of components is critical for the success of our method on complex dexterous manipulation tasks and allows it to substantially outperform these prior model-based algorithms as well as strong model-free baselines.


\vspace{-2mm}
\subsection{Model-Based Reinforcement Learning}
\vspace{-2mm}

We first define the model-based RL problem. Consider a Markov decision process (MDP) with a set of states $\mathcal{S}$, a set of actions $\mathcal{A}$, and a state transition distribution $p(s'|s,a)$ describing the result of taking action $a$ from state $s$. The task is specified by a bounded reward function $r(s,a)$, and the goal of RL is to select actions in such a way as to maximize the expected sum of rewards over a trajectory. Model-based RL aims to solve this problem by first learning an approximate model $\hat p_{\theta}(s'|s,a)$, parameterized by $\theta$, that approximates the unknown transition distribution $p(s'|s,a)$ of the underlying system dynamics. The parameters $\theta$ can be learned to maximize the log-likelihood of observed data $\mathcal{D}$, and the learned model's predictions can then be used to either learn a policy or, as we do in this work, to perform online planning to select optimal actions.


\vspace{-2mm}
\subsection{Learning the Dynamics}
\vspace{-2mm}

We use a deep neural network model to represent $\hat p_{\theta}(s'|s,a)$, such that our model has enough capacity to capture the complex interactions involved in dexterous manipulation. We use a parameterization of the form $\hat p_{\theta}(s'|s,a) = \mathcal{N}(\hat{f}_\theta(s,a), \Sigma)$, where the mean $\hat{f}_\theta(s,a)$ is given by a neural network, and the covariance $\Sigma$ of the conditional Gaussian distribution can also be learned (although we found this to be unnecessary for good results). As prior work has indicated, capturing epistemic uncertainty in the network weights is indeed important in model-based RL, especially with high-capacity models that are liable to overfit to the training set and extrapolate erroneously outside of it. A simple and inexpensive way to do this is to employ bootstrap ensembles~\cite{chua2018pets}, which approximate the posterior $p(\theta | \mathcal{D})$ with a set of $E$ models, each with parameters $\theta_i$. For deep models, prior work has observed that bootstrap resampling is unnecessary, and it is sufficient to simply initialize each model $\theta_i$ with a different random initialization $\theta^0_i$ and use different batches of data $D_i$ at each train step~\cite{chua2018pets}. We note that this supervised learning setup makes more efficient use of the data than the counterpart model-free methods, since we get dense training signals from each state transition and we are able to use all data (even off-policy data) to make training progress.


\vspace{-2mm}
\subsection{Online Planning for Closed-Loop Control}
\vspace{-2mm}

In our method, we use online planning with MPC to select actions via our model predictions. At each time step, our method performs a short-horizon trajectory optimization, using the model to predict the outcomes of different action sequences. We can use a variety of gradient-free optimizers to address this optimization; we describe a few particular choices below, each of which builds on the previous ones, with the last and most sophisticated optimizer being the one that is used by our \ourmethod ~algorithm.

\vspace{-3mm}
\paragraph{Random Shooting:}

The simplest gradient-free optimizer simply generates $N$ independent random action sequences $\{A_0 \dots A_N\}$, where each sequence $A_i=\{a^i_0 \dots a^i_{H-1}\}$ is of length $H$ action. Given a reward function $r(s,a)$ that defines the task, and given future state predictions $\hat{s}_{t+1} = f_{\theta}(\hat{s}_t, a_t) + \hat{s}_t$ from the learned dynamics model $f_{\theta}$, the optimal action sequence $A_{i^*}$ is selected to be the one corresponding to the sequence with highest predicted reward: $i^* = \arg\max_i R_i = \arg\max_i \sum_{t'=t}^{t+H-1} r(\hat{s}_{t'}, a_{t'}^i).$ This approach has been shown to achieve success on continuous control tasks with learned models~\citep{nagabandi2018mbrl}, but it has numerous drawbacks: it scales poorly with the dimension of both the planning horizon and the action space, and it often is insufficient for achieving high task performance since a sequence of actions sampled at random often does not directly lead to meaningful behavior.

\vspace{-3mm}
\paragraph{Iterative Random-Shooting with Refinement:} 

To address these issues, much prior work~\citep{botev2013cross} has instead taken a cross-entropy method (CEM) approach, which begins as the random shooting approach, but then does this sampling for multiple iterations $m \in \{0 \dots M\}$ at each time step. The top $J$ highest-scoring action sequences from each iteration are used to update and refine the mean and variance of the sampling distribution for the next iteration, as follows:
\begin{align}
A_i &= \{ a_0^i \dots a_{H-1}^i \}, ~\text{where} ~~a_t^i \sim \mathcal{N}(\mu_t^m,\Sigma_t^m) ~~\forall i \in N, t \in {0 \dots H-1} \nonumber \\
A_\text{elites} &= \text{sort}(A_i)[-J:] \nonumber \\
\mu_t^{m+1} &= \alpha * \text{mean}(A_\text{elites}) + (1-\alpha)\mu_t^m ~~\forall t  \in {0 \dots H-1} \nonumber \\
\Sigma_t^{m+1} &= \alpha * \text{var}(A_\text{elites}) + (1-\alpha)\Sigma_t^m ~~\forall t  \in {0 \dots H-1}
\end{align}
After $M$ iterations, the optimal actions are selected to be the resulting mean of the action distribution.

\vspace{-3mm}
\paragraph{Filtering and Reward-Weighted Refinement:}

While CEM is a stronger optimizer than random shooting, it still scales poorly with dimensionality and is hard to apply when both coordination and precision are required. \ourmethod ~instead uses a stronger optimizer that considers covariances between time steps and uses a softer update rule that more effectively integrates a larger number of samples into the distribution update. As derived by recent model-predictive path integral work~\citep{williams15mppi,lowrey2018plan}, this general update rule takes the following form for time step $t$, reward-weighting factor $\gamma$, and reward $R_k$ from each of the $N$ predicted trajectories:
\begin{align}
\mu_t = \frac{\sum_{k=0}^N (e^{\gamma \cdot R_k})(a_t^{(k)})}{\sum_{j=0}^N e^{\gamma \cdot (R_j)}} ~~\forall t \in \{0 \dots H-1 \}. \label{eqn:mppi_update}
\end{align}
Rather than sampling our action samples from a random policy or from iteratively refined Gaussians, we instead apply a filtering technique to explicitly produce smoother candidate action sequences. Given the iteratively updating mean distribution $\mu_t$ from above, we generate $N$ action sequences $a_t^i = n_t^i + \mu_t$, where each noise sample $n_t^i$ is generated using filtering coefficient $\beta$ as follows: 
\begin{align}
u_t^i &\sim \mathcal{N}(0,\Sigma) ~~\forall i \in \{0 \dots N-1\}, t \in \{0 \dots H-1\}\\
n_t^i &= \beta \cdot u_t^i + (1-\beta) \cdot n^i_{t-1} ~~\text{where} ~~n_{t<0}=0
\end{align}
By coupling time steps to each other, this filtering also reduces the effective degrees of freedom or dimensionality of the search space, thus allowing for better scaling with dimensionality.


\vspace{-2mm}
\subsection{Overview}
\vspace{-2mm}

Each of the optimization methods described in the previous sections can be combined with our ensemble-based dynamics models to produce a complete model-based RL method. In our approach, the reward $R_k$ for each action sequence is calculated to be the mean predicted reward across all models $\theta_i$ of the ensemble, allowing for model disagreement to affect our chosen actions. After using the predicted rewards to optimize for an $H$-step candidate action sequence, we employ MPC, where the agent executes only the first action $a_t^{i^*}$, receives updated state information $s_{t+1}$, and then replans at the following time step. The closed-loop method of replanning using updated information at every time step helps to mitigate some model inaccuraries by preventing accumulating model error. Note this control procedure also allows us to easily swap out new reward functions or goals at run-time, independent of the trained model. Overall, the full procedure of \ourmethod ~involves iteratively performing actions in the real world (through online planning with the use of a learned model), and then using those observations to update that learned model.

%% file: 05_results.tex
\vspace{-2mm}
\section{Experimental Evaluation} 
\label{sec:results}
\vspace{-2mm}

Our evaluation aims to address the following questions: (1) Can our method autonomously learn to accomplish a variety of complex dexterous manipulation tasks? (2) What is the effect of the various design decisions in our method? (3) How does our method's performance as well as sample efficiency compare to that of other state-of-the-art algorithms? (4) How general and versatile is the learned model? (5) Can we apply these lessons learned from simulation to enable a 24-DoF humanoid hand to manipulate free-floating objects in the real world?

\vspace{-2mm}
\subsection{Task Suite}


\begin{wrapfigure}{r}{0.5\columnwidth}
\vspace{-11mm}
\centering
\includegraphics[width=0.15\textwidth,trim={0 0 0 5cm},clip]{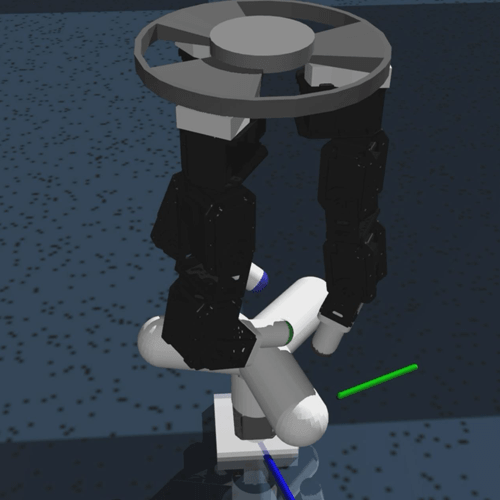}
\includegraphics[width=0.15\textwidth,trim={0 0 0 5cm},clip]{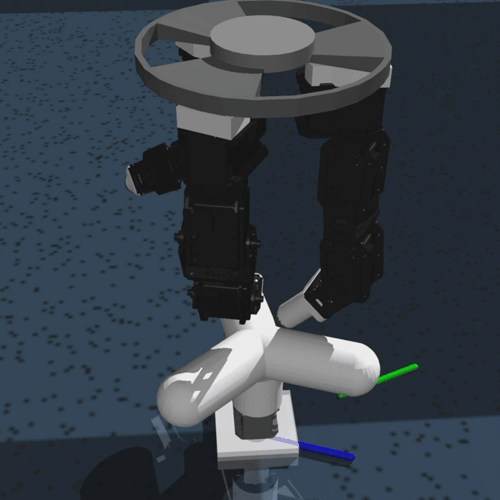}
\includegraphics[width=0.15\textwidth,trim={0 0 0 5cm},clip]{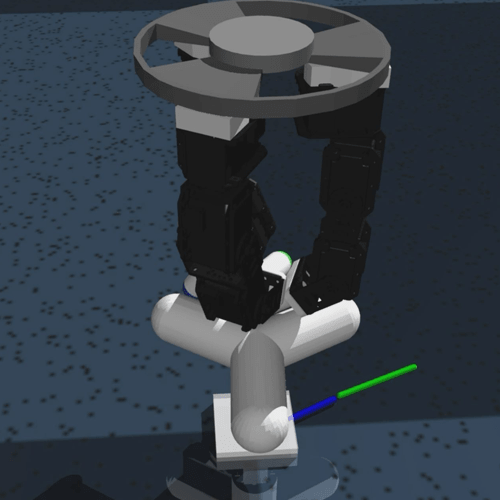}\\

\includegraphics[width=0.15\textwidth,trim={0 5cm 0 0},clip]{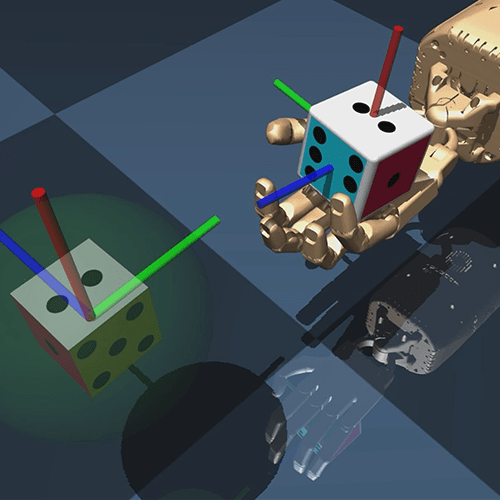}
\includegraphics[width=0.15\textwidth,trim={0 5cm 0 0},clip]{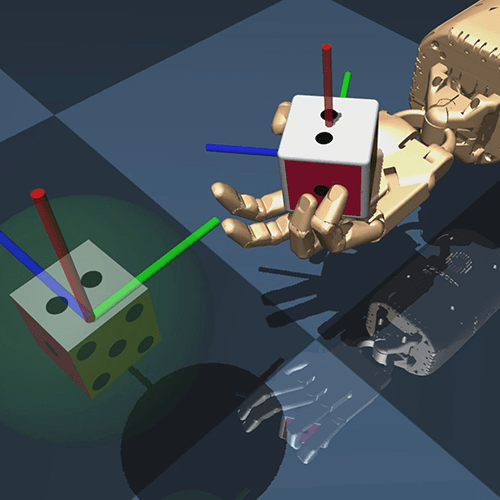}
\includegraphics[width=0.15\textwidth,trim={0 5cm 0 0},clip]{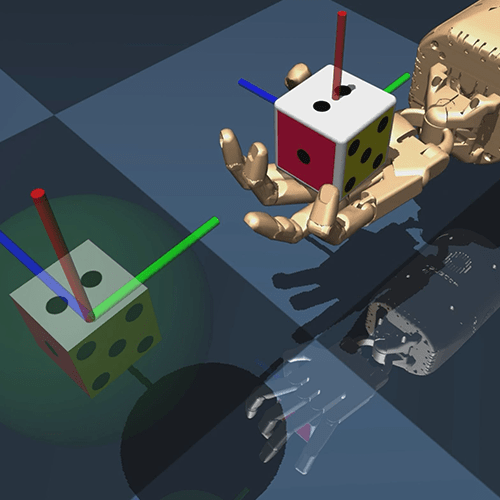}\\

\includegraphics[width=0.15\textwidth,trim={0 3cm 0 0},clip]{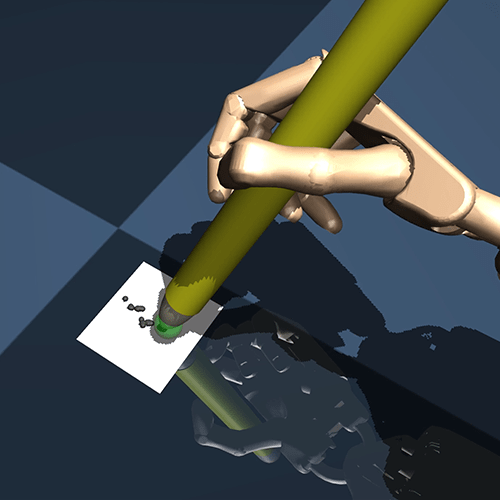}
\includegraphics[width=0.15\textwidth,trim={0 3cm 0 0},clip]{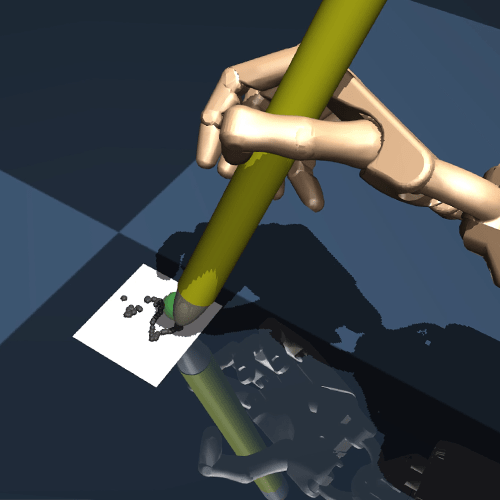}
\includegraphics[width=0.15\textwidth,trim={0 3cm 0 0},clip]{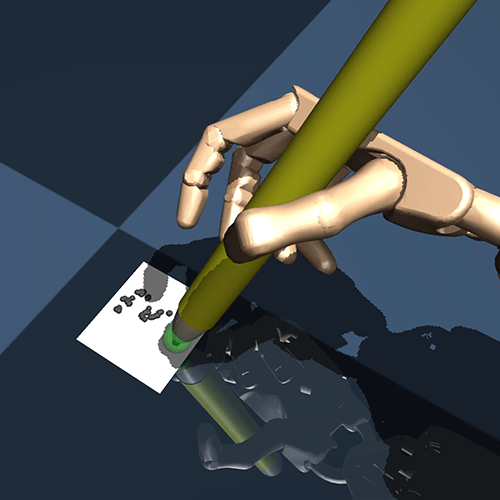}\\

\includegraphics[width=0.15\textwidth,trim={0 0 0 2cm},clip]{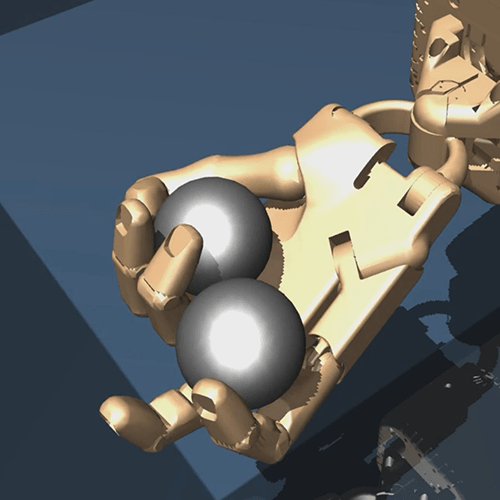}
\includegraphics[width=0.15\textwidth,trim={0 0 0 2cm},clip]{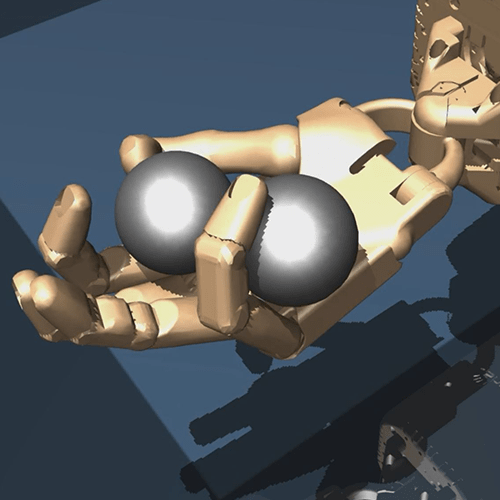}
\includegraphics[width=0.15\textwidth,trim={0 0 0 2cm},clip]{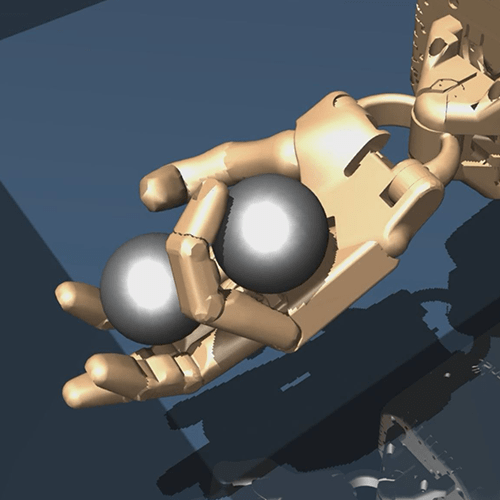}\\

\includegraphics[width=0.15\textwidth,trim={0 0 0 2.5cm},clip]{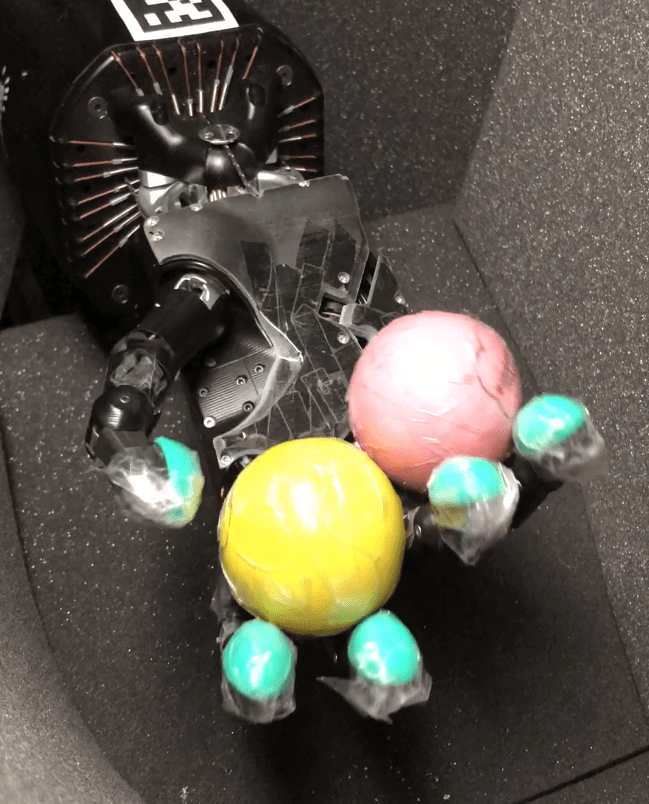}
\includegraphics[width=0.15\textwidth,trim={0 0 0 2.5cm},clip]{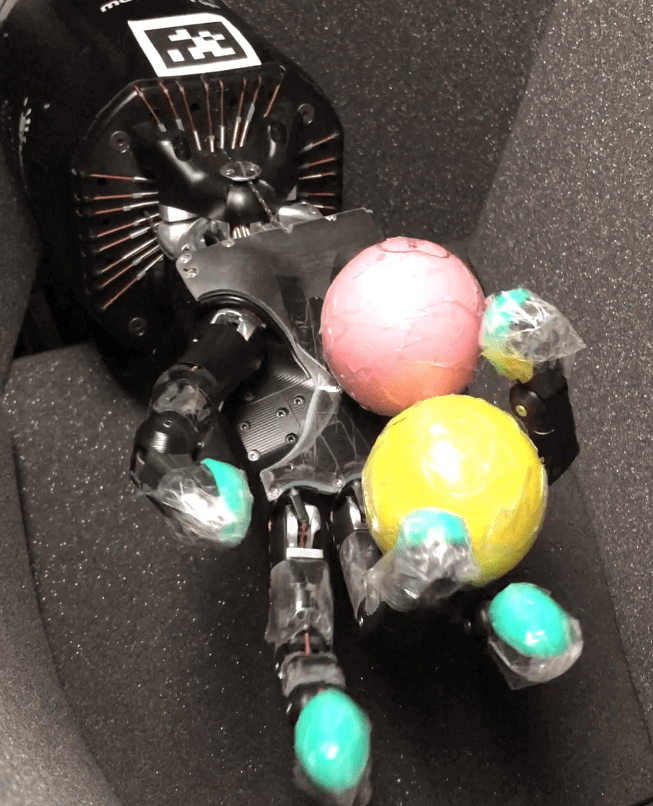}
\includegraphics[width=0.15\textwidth,trim={0 0 0 2.5cm},clip]{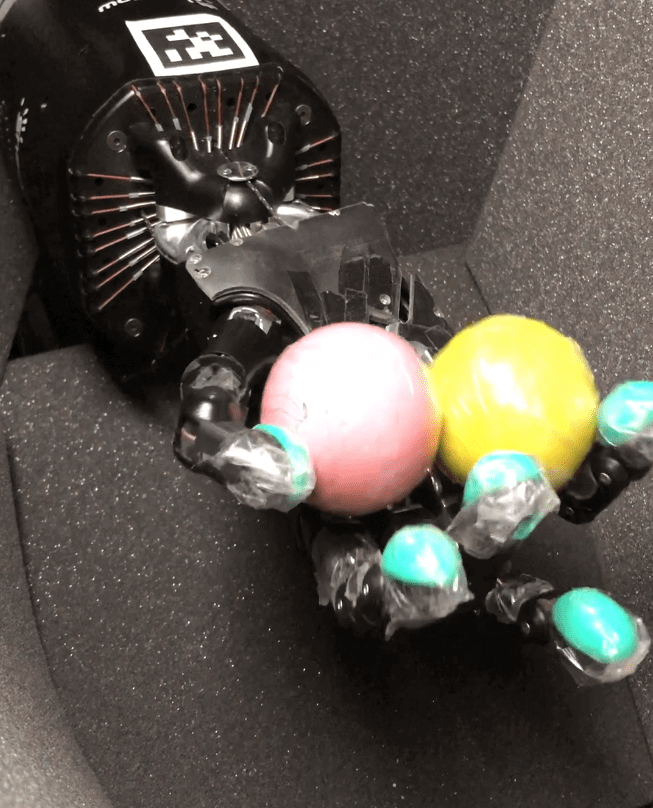}\\
\vspace{-2mm}
\caption{\footnotesize{Successful executions of \ourmethod ~on simulated and real-world dexterous manipulation tasks.}}
\vspace{-0.2in}
\label{fig:qual}
\end{wrapfigure}


Some of the main challenges in dexterous manipulation involve the high dimensionality of the hand, the prevalence of complex contact dynamics that must be utilized and balanced to manipulate free floating objects, and the potential for failure. We identified a set of tasks (\autoref{fig:tasks}) that specifically highlight these challenges by requiring delicate, precise, and coordinated movement. With the final goal in mind of real-world experiments on a 24-DoF hand, we first set up these selected dexterous manipulation tasks in MuJoCo~\cite{mujoco} and conducted experiments in simulation on two robotic platforms: a 9-DoF three-fingered hand, and a 24-DoF five-fingered hand. We show in \autoref{fig:qual} the successful executions of \ourmethod ~on these tasks, which took between 1-2 hours worth of data for simulated tasks and 2-4 hours worth of data for the real-world experiments. We include implementation details, reward functions, and hyperparameters in the appendix, but we first provide brief task overviews below. 

\vspace{-2mm}
\paragraph{Valve Turning:}
This starter task for looking into manipulation challenges uses a 9-DoF hand (D'Claw)~\citep{zhu2019dexterous} to turn a 1-DoF valve to arbitrary target locations. Here, the fingers must come into contact with the object and coordinate themselves to make continued progress toward the target. 

\vspace{-2mm}
\paragraph{In-hand Reorientation:}
Next, we look into the manipulation of free-floating objects, where most maneuvers lead to task failures of dropping objects, and thus sharp discontinuities in the dynamics. Successful manipulation of free-floating objects, such as this task of reorienting a cube into a goal pose, requires careful stabilization strategies as well as re-grasping strategies. Note that these challenges exist even in cases where system dynamics are fully known, let alone with learned models. 

\vspace{-2mm}
\paragraph{Handwriting:}
In addition to free-floating objects, the requirement of precision is a further challenge when planning with approximate models. Handwriting, in particular, requires precise maneuvering of all joints in order to control the movement of the pencil tip and enable legible writing. Furthermore, the previously mentioned challenges of local minima (i.e., holding the pencil still), abundant terminal states (i.e., dropping the pencil), and the need for simultaneous coordination of numerous joints still apply. This task can also test flexibility of learned behaviors, by requiring the agent to follow arbitrary writing patterns as opposed to only a specific stroke. 

\vspace{-2mm}
\paragraph{Baoding Balls:}
While manipulation of one free object is already a challenge, sharing the compact workspace with other objects exacerbates the challenge and truly tests the dexterity of the manipulator. We examine this challenge with Baoding balls, where the goal is to rotate two balls around the palm without dropping them. The objects influence the dynamics of not only the hand, but also each other; inconsistent movement of either one knocks the other out of the hand, leading to failure. 

\vspace{-2mm}
\subsection{Ablations and Analysis of Design Decisions}
\vspace{-2mm}

\begin{figure}[b]
\label{ablations}
\vspace{-0.1in}
\centering
\includegraphics[width=0.3\columnwidth]{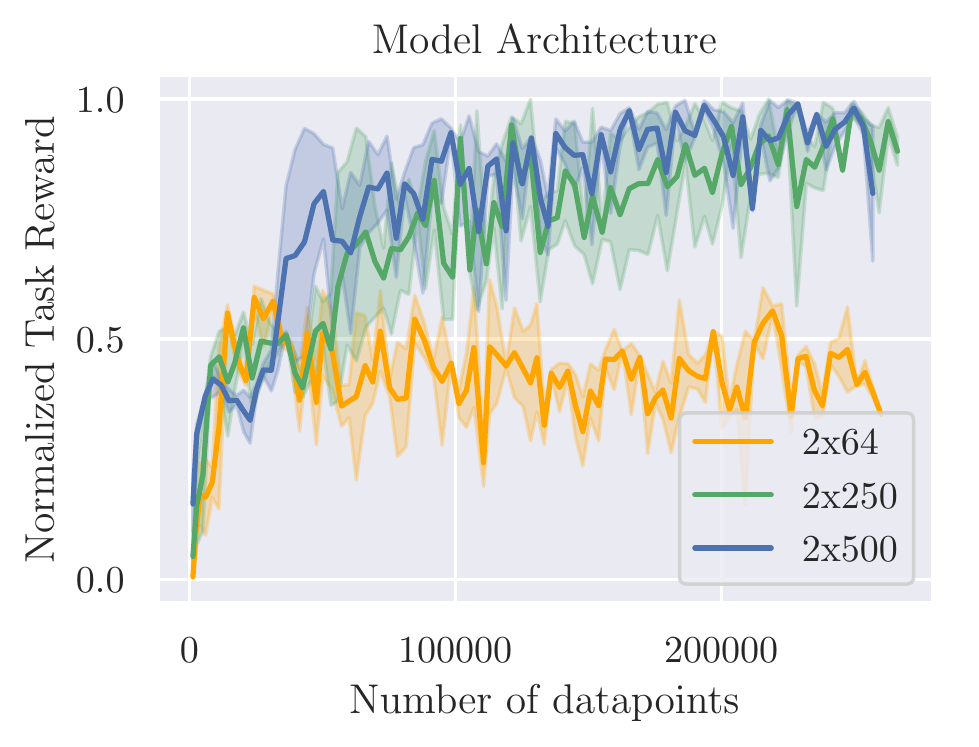}
\includegraphics[width=0.3\columnwidth]{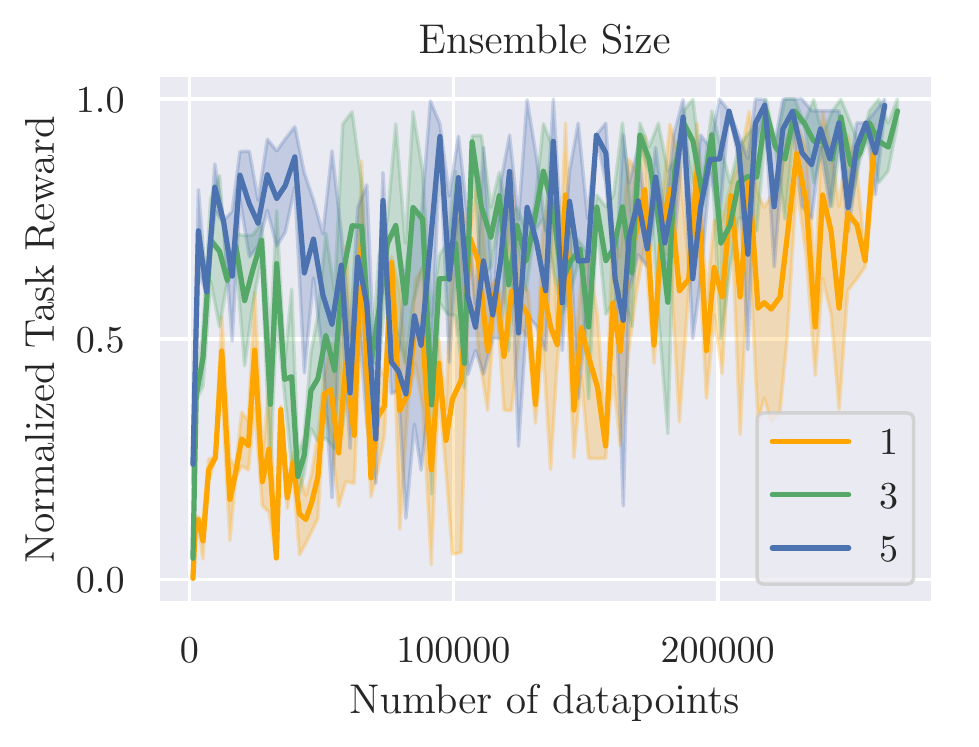}
\includegraphics[width=0.3\columnwidth]{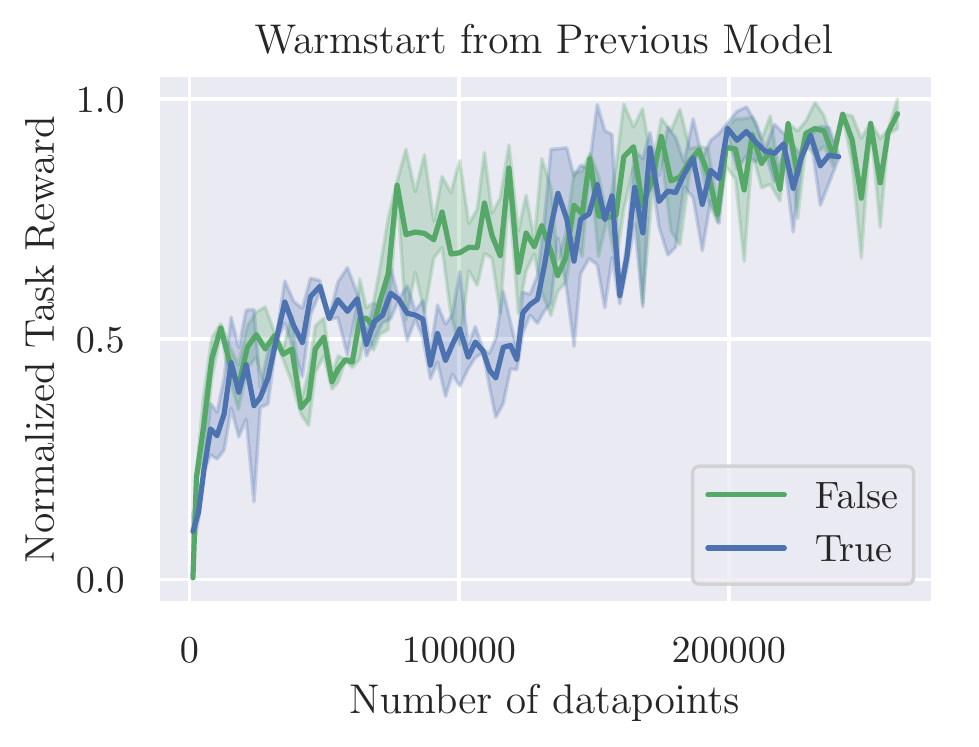}
\includegraphics[width=0.3\columnwidth]{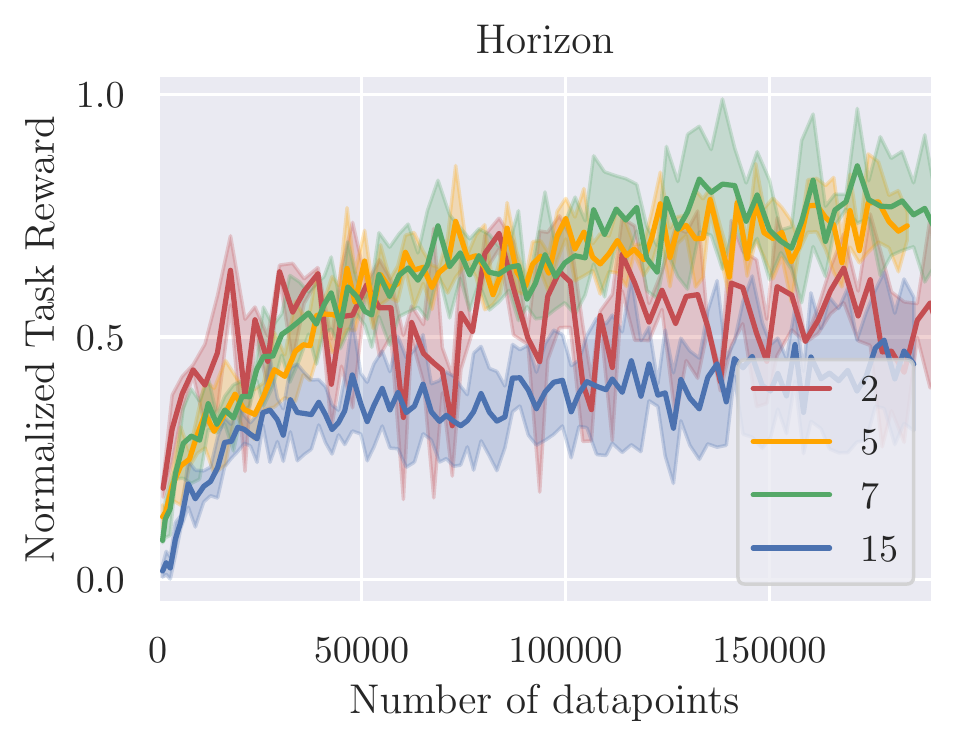}
\includegraphics[width=0.3\columnwidth]{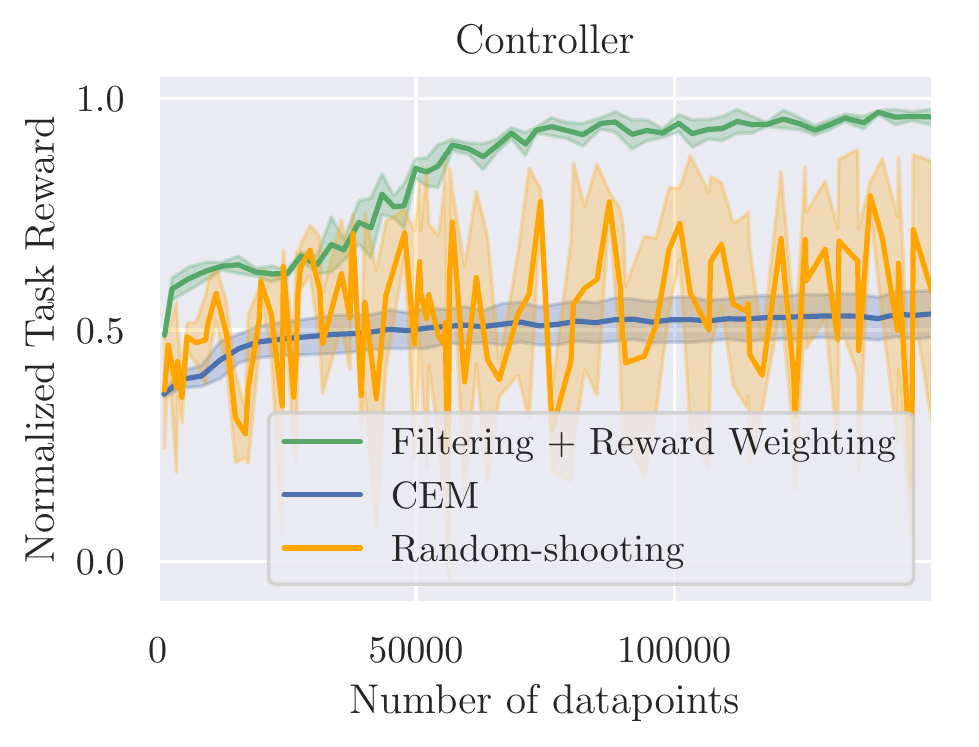}
\includegraphics[width=0.3\columnwidth]{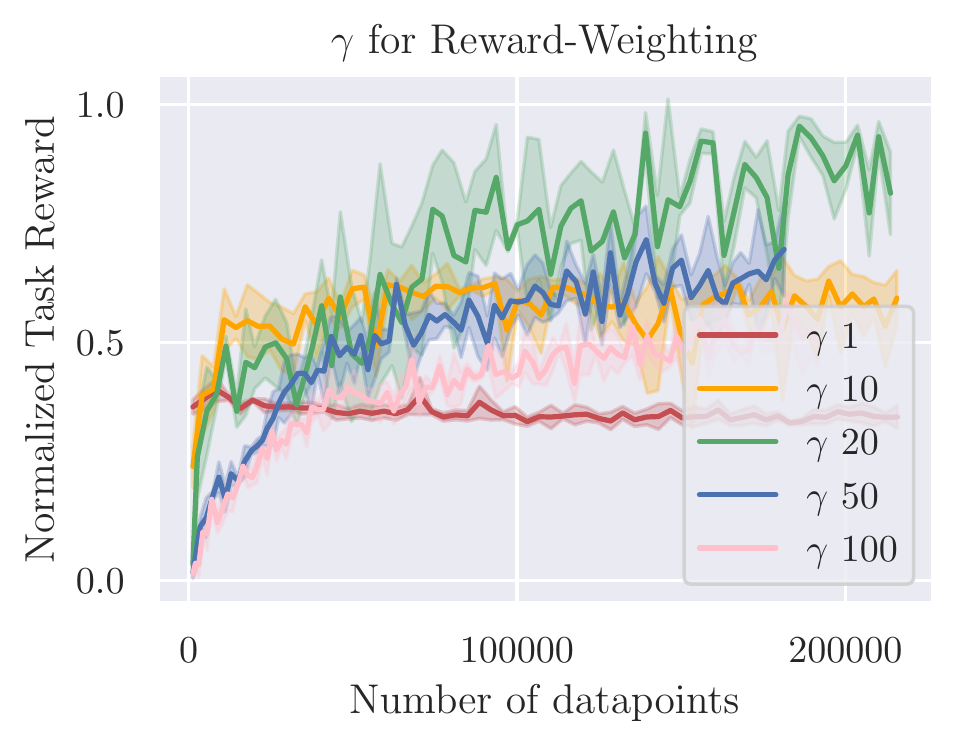}
\vspace{-2mm}
\caption{\footnotesize{Baoding task performance for various design decisions: (Top) model architecture, ensemble size, warmstarting model weights, and (Bottom) planning horizon, controller type, and reward-weighting variable $\gamma$.}}
\label{fig:ablations}
\vspace{-0.15in}
\end{figure}

In our first set of experiments (Fig.~\ref{fig:ablations}), we evaluate the impact of the design decisions for our model and our online planning method. We use the Baoding balls task for these experiments, though we observed similar trends on other tasks. In the first plot, we see that a sufficiently large architecture is crucial, indicating that the model must have enough capacity to represent the complex dynamical system. In the second plot, we see that the use of ensembles is helpful, especially earlier in training when non-ensembled models can overfit badly and thus exhibit overconfident and harmful behavior. This suggests that ensembles are an enabling factor in using sufficiently high-capacity models. In the third plot, we see that there is not much difference between resetting model weights randomly at each training iteration versus warmstarting them from their previous values.

In the fourth plot, we see that using a planning horizon that is either too long or too short can be detrimental: Short horizons lead to greedy planning, while long horizons suffer from compounding errors in the predictions. In the fifth plot, we study the type of planning algorithm and see that \ourmethod, with action smoothing and soft updates, greatly outperforms the others. In the final plot, we study the effect of the $\gamma$ reward-weighting variable, showing that medium values provide the best balance of dimensionality reduction and smooth integration of action samples versus loss of control authority. Here, too soft of a weighting leads to minimal movement of the hand, and too hard of a weighting leads to aggressive behaviors that frequently drop the objects.

\vspace{-2mm}
\subsection{Comparisons}
\vspace{-2mm}


\begin{figure}
\centering
\begin{minipage}{.3\textwidth}
\includegraphics[height=3.4cm]{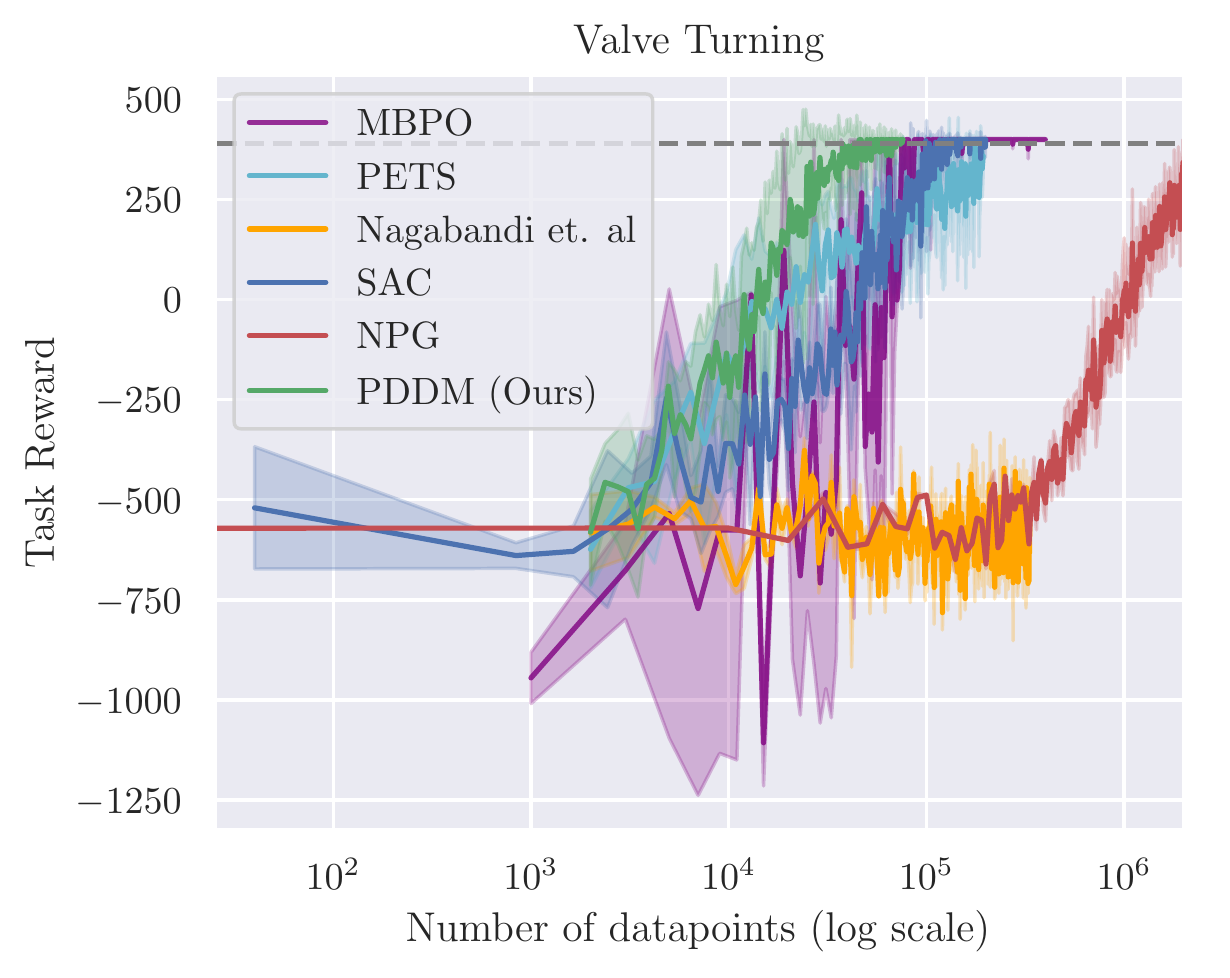}
\vspace{-5mm}
\caption{\footnotesize{Most methods learn this easier task of valve turning, though our method learns fastest.}}
\label{fig:valve}
\end{minipage} \hspace{0.1in}
\begin{minipage}{.65\textwidth}
\includegraphics[height=3.4cm]{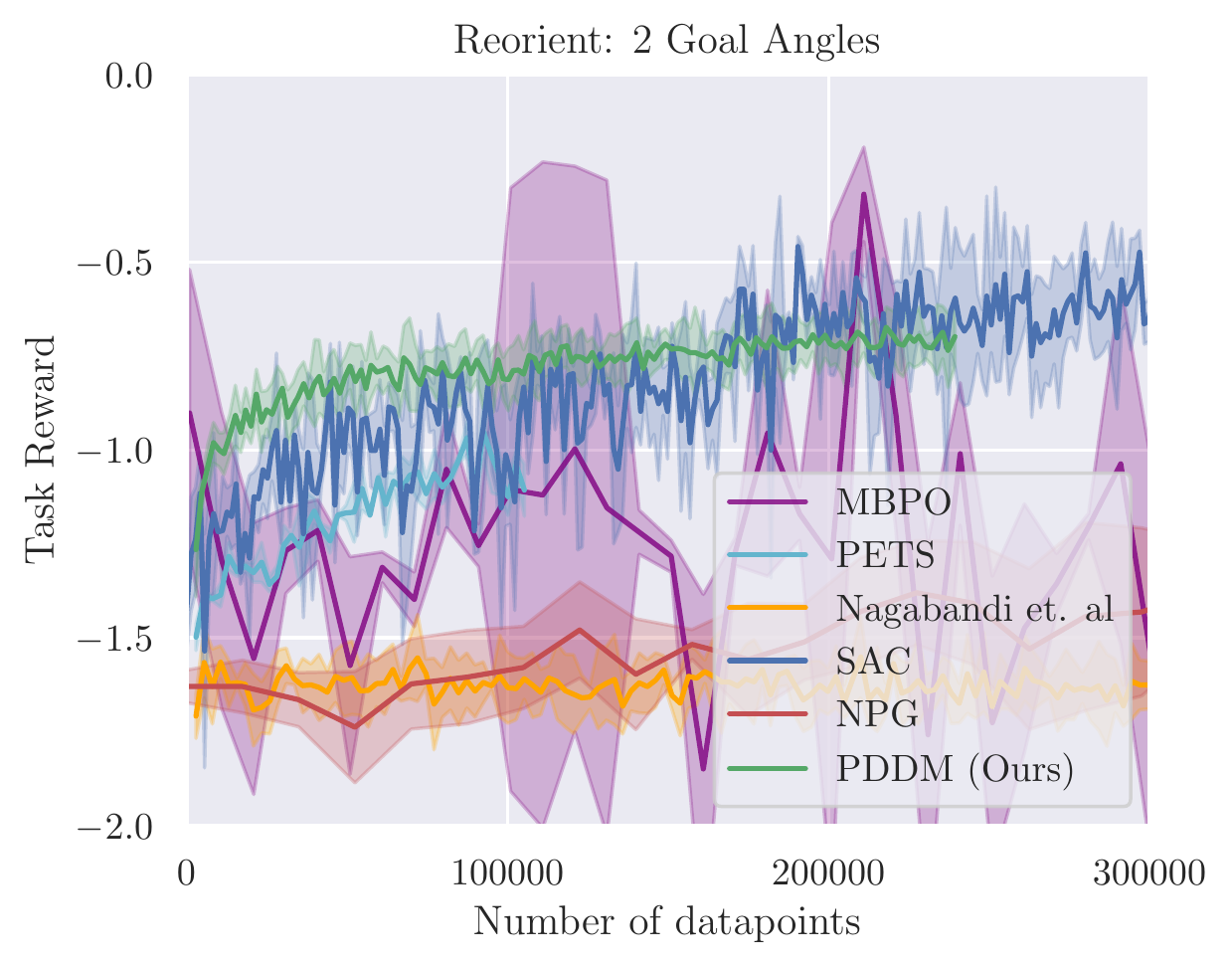}
\includegraphics[height=3.4cm]{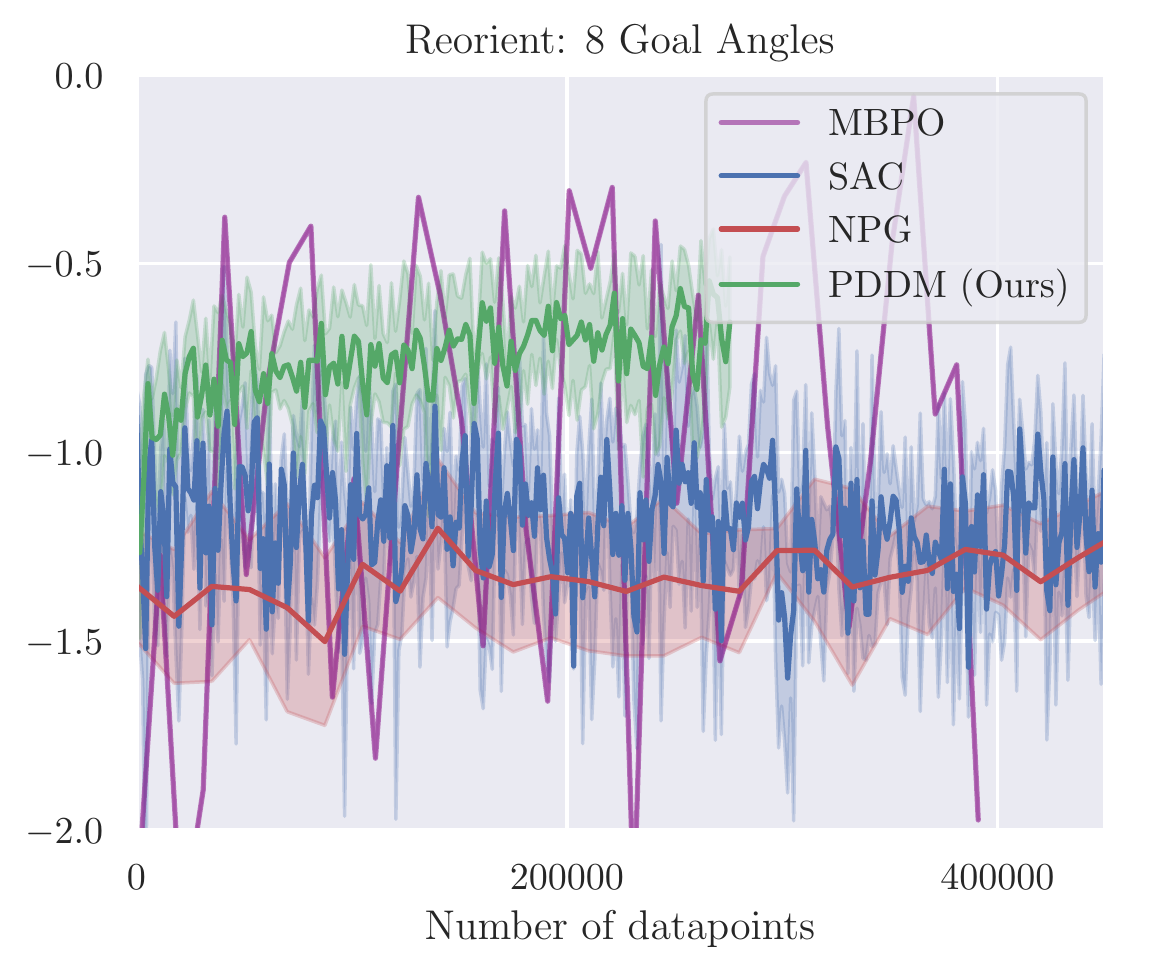}
\vspace{-2mm}
\caption{\footnotesize{In-hand reorientation of a cube. Our method achieves the best results with 2 and 8 goal angles, and model-free algorithms and methods that directly learn a policy, such as MBPO, struggle with 8 goal angles.}}
\label{fig:cube}
\end{minipage}
\end{figure}

\begin{figure}
\centering

\begin{minipage}{.65\textwidth}
\includegraphics[height=3.2cm]{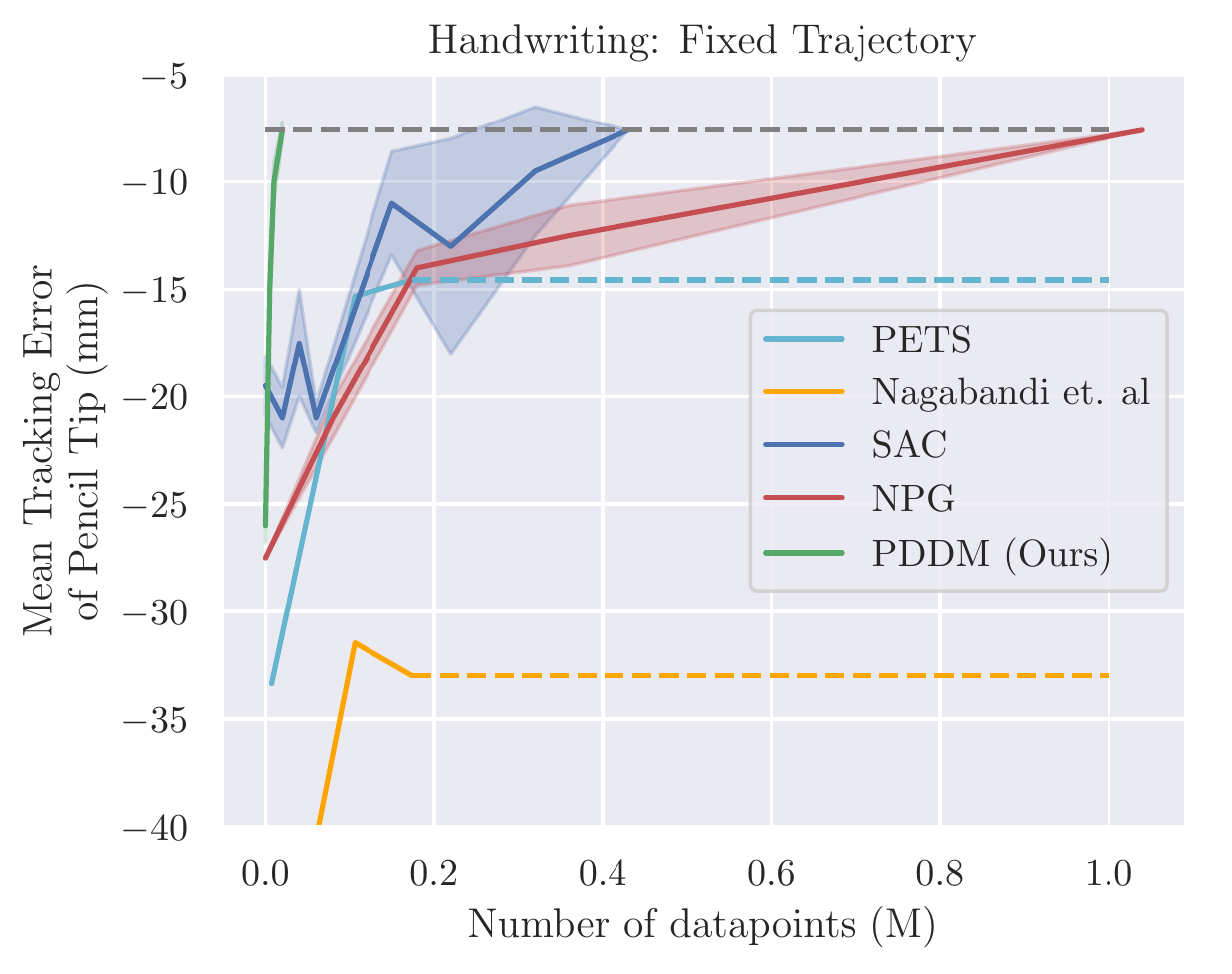}
\includegraphics[height=3.2cm]{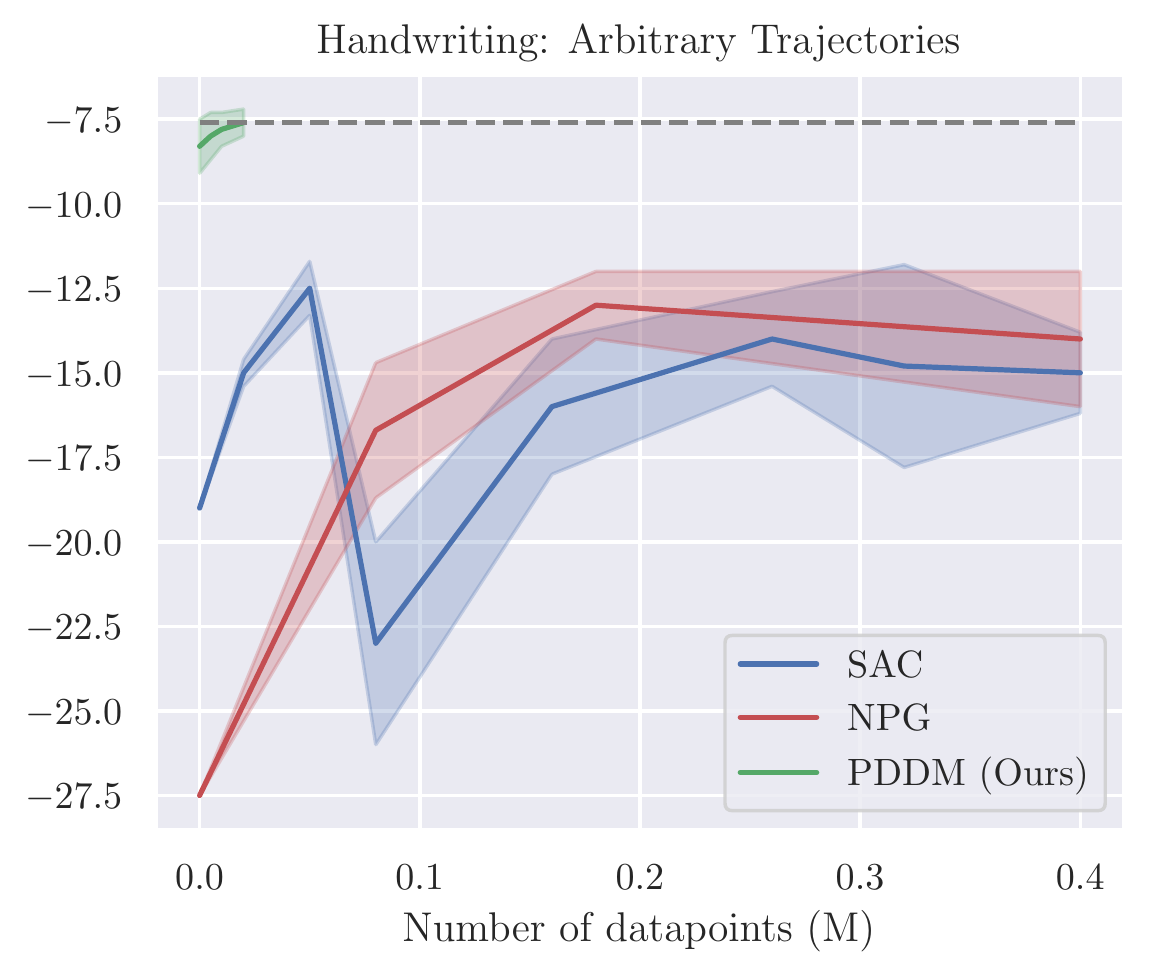}
\vspace{-2mm}
\caption{\footnotesize{Manipulating a pencil to make its tip follow (left) a fixed desired path, where \ourmethod ~learns substantially faster than SAC and NPG, and (right) arbitrary paths, where only \ourmethod ~succeeds.}}
\label{fig:handwriting} 
\end{minipage} \hspace{0.05in}
\begin{minipage}{.3\textwidth} 
\includegraphics[height=3.2cm]{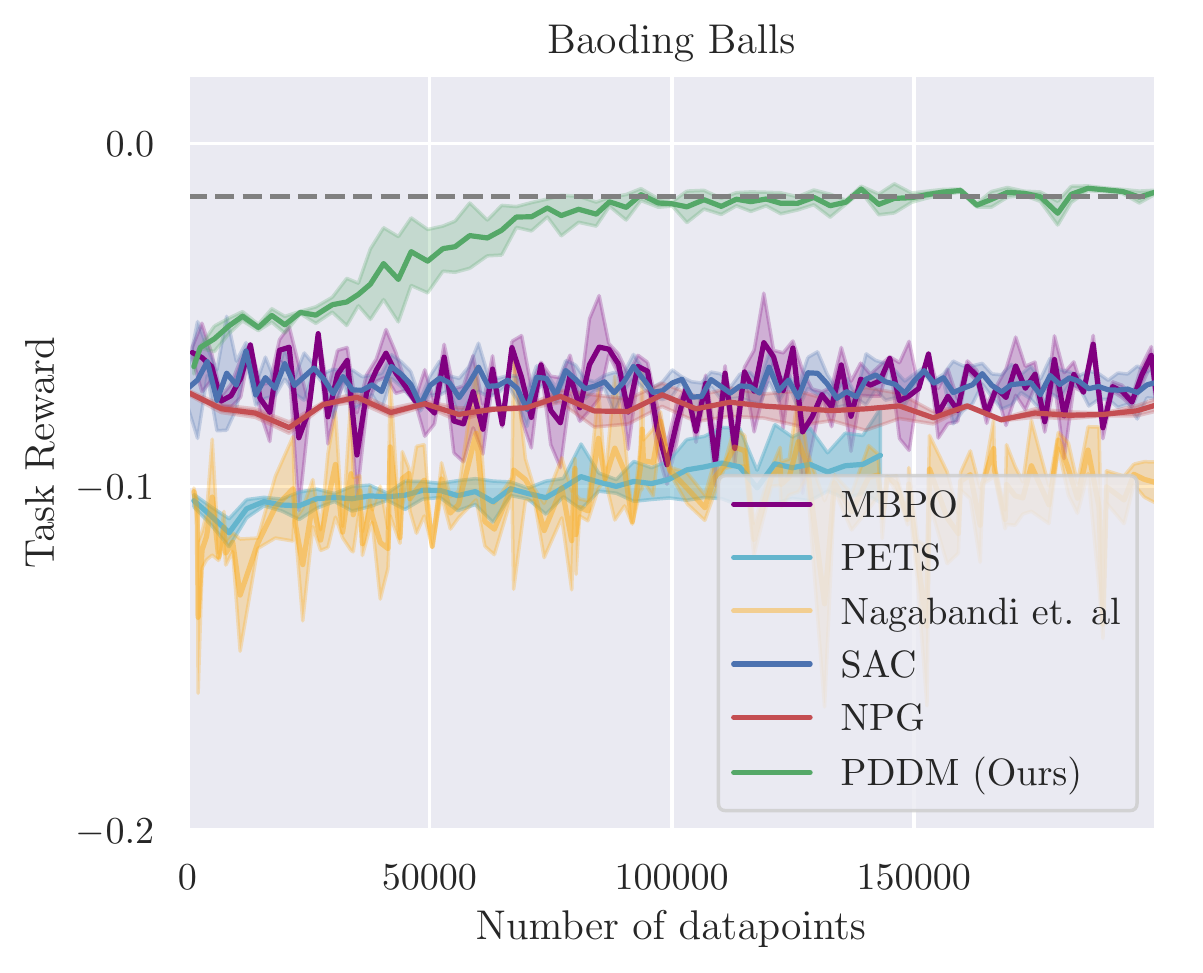}
\vspace{-2mm}
\caption{\footnotesize{\ourmethod ~outperforms prior model-based and model-free methods.}}
\label{fig:baoding}
\end{minipage}
\vspace{-4mm}

\end{figure}


In this section, we compare our method to the following state-of-the-art model-based and model-free RL algorithms: \textbf{Nagabandi et. al}~\citep{nagabandi2018mbrl} learns a deterministic neural network model, combined with a random shooting MPC controller; \textbf{PETS}~\citep{kurutach2018model} combines uncertainty-aware deep network dynamics models with sampling-based uncertainty propagation; \textbf{NPG}~\citep{kakade2002natural} is a model-free natural policy gradient method, and has been used in prior work on learning manipulation skills~\citep{rajeswaran2017learning}; \textbf{SAC}~\citep{haarnoja2018soft} is an off-policy model-free RL algorithm; \textbf{MBPO}~\citep{janner2019trust} is a recent hybrid approach that uses data from its model to accelerate policy learning. On our suite of dexterous manipulation tasks, \ourmethod ~consistently outperforms prior methods both in terms of learning speed and final performance, even solving tasks that prior methods cannot.


\vspace{-0.15in}
\paragraph{Valve turning:} We first experiment with a three-fingered hand rotating a valve, with starting and goal positions chosen randomly from the range [$-\pi, \pi$]. On this simpler task, we confirm that most of the prior methods do in fact succeed, and we also see that even on this simpler task, policy gradient approaches such as NPG require prohibitively large amounts of data (note the log scale of~\autoref{fig:valve}).


\vspace{-0.15in}
\paragraph{In-hand reorientation:} Next, we scale up our method to a 24-DoF five-fingered hand reorienting a free-floating object to arbitrary goal configurations (\autoref{fig:cube}). First, we prescribe two possible goals of either left or right 90$\degree$ rotations; here, SAC behaves similarly to our method (actually attaining a higher final reward), and NPG is slow as expected, but does achieve the same high reward after $6e6$ steps. However, when we increase the number of possible goals to 8 different options (90$\degree$ and 45$\degree$ rotations in the left, right, up, and down directions), we see that our method still succeeds, but the model-free approaches get stuck in local optima and are unable to fully achieve even the previously attainable goals. This inability to effectively address a ``multi-task" or ``multi-goal" setup is indeed a known drawback for model-free approaches, and it is particularly pronounced in such goal-conditioned tasks that require flexibility. These additional goals do not make the task harder for \ourmethod, because even in learning 90$\degree$ rotations, it is building a model of its interactions rather than specifically learning to get to those angles.


\vspace{-0.15in}
\paragraph{Handwriting:} To further scale up task complexity, we experiment with handwriting, where the base of the hand is fixed and all writing must be done through coordinated movement of the fingers and the wrist. We perform two variations of this task: (a) the agent is trained and tested on writing a single fixed trajectory, and (b) the agent is trained with the goal of following arbitrary trajectories, but is evaluated on the fixed trajectory from (a). Although even a fixed writing trajectory is challenging, writing arbitrary trajectories requires a degree of flexibility that is exceptionally difficult for prior methods. We see in~\autoref{fig:handwriting} that prior model-based approaches don't actually solve this task (values below the grey line correspond to holding the pencil still near the middle of the paper). Our method, SAC, and NPG solve the task for a single fixed trajectory, but the model-free methods fail when presented with arbitrary trajectories and become stuck in a local optima when trying to write arbitrary trajectories. Even SAC, which has a higher entropy action distribution and therefore achieves better exploration, is unable to extract the finer underlying skill due to the landscape for successful behaviors being quite narrow.


\vspace{-0.15in}
\paragraph{Baoding Balls:} This task is particularly challenging due to the inter-object interactions, which can lead to drastically discontinuous dynamics and frequent failures from dropping the objects. We were unable to get the other model-based or model-free methods to succeed at this task (\autoref{fig:baoding}), but \ourmethod ~solves it using just 100,000 data points, or 2.7 hours worth of data. Additionally, we can employ the model that was trained on 100-step rollouts to then run for much longer (1000 steps) at test time. The model learned for this task can also be repurposed, without additional training, to perform a variety of related tasks (see video): moving a single ball to a goal location in the hand, posing the hand, and performing clockwise rotations instead of the learned counter-clockwise ones.


\vspace{-2mm}
\subsection{Learning Real-World Manipulation of Baoding Balls}
\label{sec:shadowhand}
\vspace{-2mm}

Finally, we present an evaluation of \ourmethod ~on a real-world anthropomorphic robotic hand. We use the 24-DoF Shadow Hand to manipulate Baoding balls, and we train our method entirely with real-world experience, without any simulation or prior knowledge of the system.

\begin{figure}[t]
\vspace{-0.1in}
\centering
\includegraphics[height=0.2\columnwidth,trim={1.5cm 0 0 0},clip]{figs/shadowhand_baoding.png}
\includegraphics[height=0.2\columnwidth]{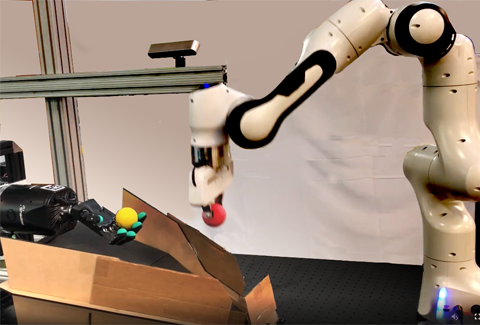}
\includegraphics[height=0.2\columnwidth]{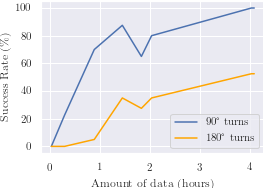}
\vspace{-2mm}
\caption{\footnotesize{Real-world Baoding balls hardware setup with the ShadowHand (left), Franka-Emika arm used for the automated reset mechanism (middle), and the resulting success rates for $90\degree$ and $180\degree$ turns (right).}}
\label{fig:hardware_baoding_results}
\vspace{-0.1in}
\end{figure}

\textbf{Hardware setup:} In order to run this experiment in the real world (\autoref{fig:hardware_baoding_results}), we use a camera tracker to produce 3D position estimates for the Baoding balls. We employ a dilated CNN model for object tracking, modeled after KeypointNet~\citep{Suwajanakorn:2018}, which takes a 280x180 RGB stereo image pair as input from a calibrated 12 cm baseline camera rig. Additional details on the tracker are provided in Appendix A. As shown in the supplementary video, we also implement an automated reset mechanism, which consists of a ramp that funnels the dropped Baoding balls to a specific position and then triggers a pre-preprogrammed 7-DoF Franka-Emika arm to use its parallel jaw gripper to pick them up and return them to the Shadow Hand's palm. The planner commands the hand at 10Hz, which is communicated via a 0-order hold to the low-level position controller that operates at 1Khz. The episode terminates if the specific task horizon of 10 seconds has elapsed or if the hand drops either ball, at which point a reset request is issued again. Numerous sources of delays in real robotic systems, in addition to the underactuated nature of the real Shadow Hand, make the task quite challenging in the real world.

\textbf{Results:} Our method is able to learn $90\degree$ rotations without dropping the two Baoding balls after under 2 hours of real-world training, with a success rate of about $100\%$, and can achieve a success rate of about $54\%$ on the challenging $180\degree$ rotation task, as shown in \autoref{fig:hardware_baoding_results}. An example trajectory of the robot rotating the Baoding balls using our method is shown in Figure~\ref{fig:teaser_baoding}, and videos on the project website \footnote{\url{https://sites.google.com/view/pddm/}} illustrate task progress through various stages of training. Qualitatively, we note that performance improves fastest during the first 1.5 hours of trainingl; after this, the system must learn the more complex transition of transferring the control of a Baoding ball from the pinky to the thumb (with a period of time in between, where the hand has only indirect control of the ball through wrist movement). These results illustrate that, although the real-world version of this task is substantially more challenging than its simulated counterpart, our method can learn to perform it with considerable proficiency using a modest amount of real-world training data.

%% file: 06_conclusion.tex
\vspace{-1mm}
\section{Discussion} 
\vspace{-1mm}
\label{sec:conclusion}

We presented a method for using deep model-based RL to learn dexterous manipulation skills with multi-fingered hands. We demonstrated results on challenging non-prehensile manipulation tasks, including controlling free-floating objects, agile finger gaits for repositioning objects in the hand, and precise control of a pencil to write user-specified strokes. As we show in our experiments, our method achieves substantially better results than prior deep model-based RL methods, and it also has advantages over model-free RL: it requires substantially less training data and results in a model that can be flexibly reused to perform a wide variety of user-specified tasks. In direct comparisons, our approach substantially outperforms state-of-the-art model-free RL methods on tasks that demand high flexibility, such as writing user-specified characters. In addition to analyzing the approach on our simulated suite of tasks using 1-2 hours worth of training data, we demonstrate \ourmethod ~on a real-world 24 DoF anthropomorphic hand, showing successful in-hand manipulation of objects using just 4 hours worth of entirely real-world interactions. Promising directions for future work include studying methods for planning at different levels of abstraction in order to succeed at sparse-reward or long-horizon tasks, as well as studying the effective integration of additional sensing modalities such as vision and touch into these models in order to better understand the world.

%% file: appendix.tex
\section*{Appendix}

\section{Method Overview}
\label{app:block_diagram}

\begin{figure}[h]
\centering
\includegraphics[width=0.7\columnwidth,trim={1.5cm 2.5cm 1.5cm 1.5cm},clip]{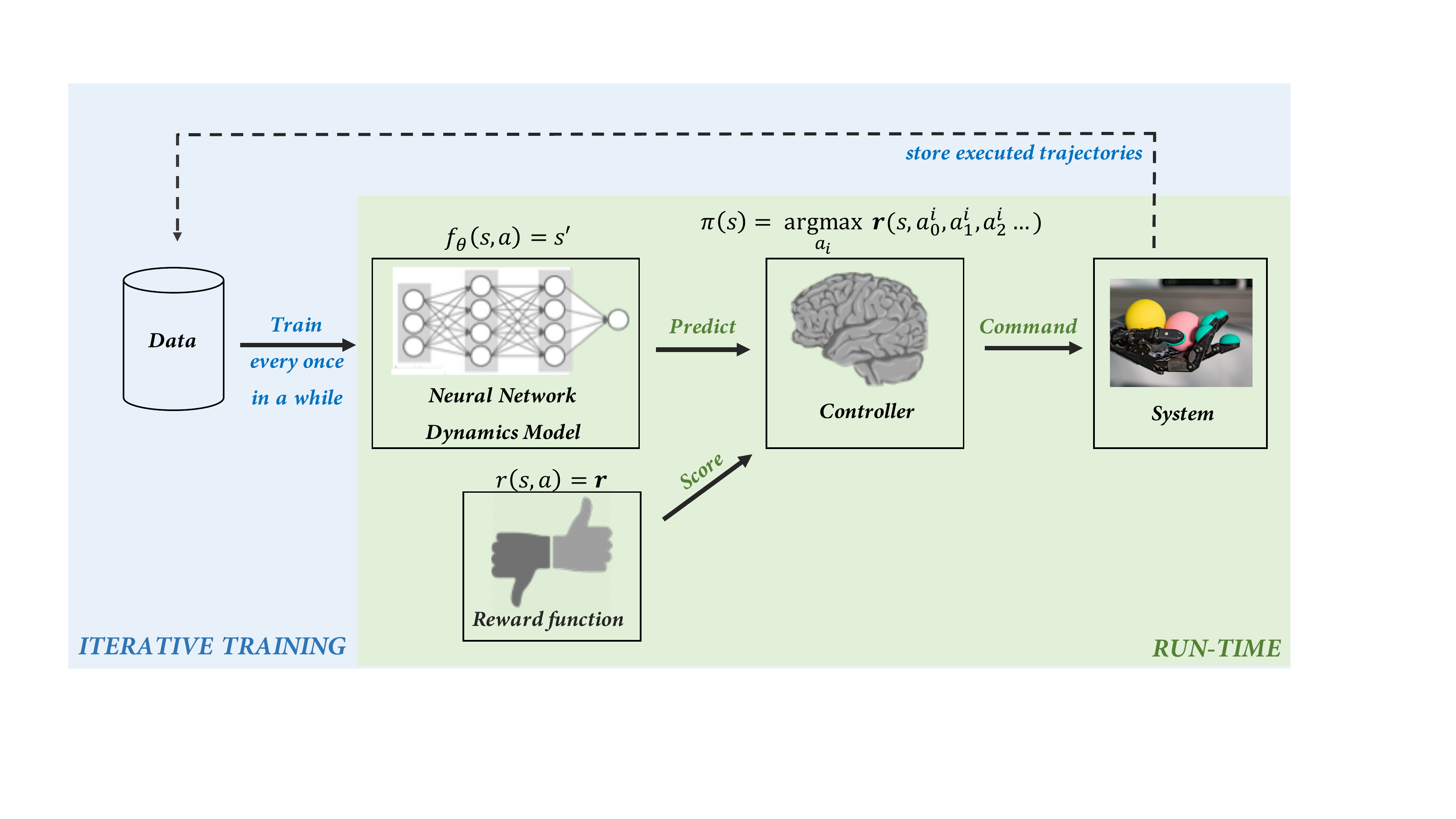}
\caption{\footnotesize{Overview diagram of our online planning with deep dynamics models (\ourmethod).}}
\label{fig:block_diagram}
\end{figure}

At a high level, this method (\autoref{fig:block_diagram}, Algorithm~\ref{alg:overview}) of online planning with deep dynamics models involves an iterative procedure of (a) running a controller to perform action selection using predictions from a trained predictive dynamics model, and (b) training a dynamics model to fit that collected data. With recent improvements in both modeling procedures as well as control schemes using these high-capacity learned models, we are able to demonstrate efficient and autonomous learning of complex dexterous manipulation tasks.

\begin{algorithm}[h]
    \caption{PDDM Overview}
    \label{alg:overview}
    \begin{algorithmic}[1]
        \STATE randomly initialize ensemble of models $\{\theta_0, \dots, \theta_M\}$
        \STATE initialize empty dataset $D \leftarrow \{\}$
        \FOR{$\text{iter}$ in range($I$)}
            \FOR{$\text{rollout}$ in range$(R)$}
                \STATE $s_0 \leftarrow$ reset env
                \FOR{$t$ in range($T$)}
                    \STATE $a \leftarrow \text{PDDM}_{\text{MPC}}(s_t, \{f_{\theta_0}, \dots, f_{\theta_M}\}, H, N, r, \gamma, \beta)$
                    \STATE $s_{t+1} \leftarrow$ take action $a$
                    \STATE $D \leftarrow (s_t, a_t, s_{t+1})$
                \ENDFOR
            \ENDFOR
            \STATE use $D$ to train models $\{f_{\theta_0}, \dots, f_{\theta_M}\}$ for $E$ epochs each
        \ENDFOR
    \end{algorithmic}
\end{algorithm}


\section{Experiment Details}
\label{app:details}

We implement the dynamics model as a neural network of 2 fully-connected hidden layers of size 500 with relu nonlinearities and a final fully-connected output layer. For all tasks and environments, we train this same model architecture with a standard mean squared error (MSE) supervised learning loss, using the Adam optimizer~\citep{Kingma2014_ICLR} with learning rate $0.001$. Each epoch of model training, as mentioned in Algorithm~\ref{alg:overview}, consists of a single pass through the dataset $D$ while taking a gradient step for every sampled batch of size 500 data points. The other relevant hyperparameters are listed in Table~\ref{table:hyper}, referenced by task. For each of these tasks, we normalize the action spaces to $[-1,1]$ and act at a control frequency of $\frac{1}{dt}$. We list reward functions and other relevant task details in Table~\ref{table:taskdetails}.

\renewcommand{\arraystretch}{2.2}
\begin{table}[h]
\caption{Hyperparameters}\label{table:hyper}
\centering
\begin{tabular}{|c|c|c|c|c|c|c|c|c|}
\hline  & $R$ & $T$ & $H$ & $N$ & $\gamma$ & $\beta$ & $M$ & $E$ \\
\hline Valve Turning & 20 & 200 & 7 & 200 & 10 & 0.6 & 3 & 40\\
\hline In-hand Reorientation & 30 & 100 & 7 & 700 & 50 & 0.7 & 3 & 40\\
\hline Handwriting & 40 & 100 & 7 & 700 & 0.5 & 0.5 & 3 & 40\\
\hline Baoding Balls & 30 & 100 & 7 & 700 & 20 & 0.7 & 3 & 40\\
\hline
\end{tabular}
\end{table}

\renewcommand{\arraystretch}{2.2}
\begin{table}[h]
\caption{Task details}\label{table:taskdetails}
\centering
\begin{tabular}{|c|c|c|c|c|}
\hline  & $dt$ (sec)  & Dim of $s$ & Dim of $a$ & Reward $r(s,a)$ \\

\hline \multirow{3}{*}{Valve Turning} & 0.15 & 21 & 9 & $-10|\text{valve}_\theta-\text{target}_\theta|$ \\
& & & & $+\mathbbm{1}{(|\text{valve}_\theta-\text{target}_\theta|<0.25)}$ \\ 
& & & & $+10 * ~\mathbbm{1}{(|\text{valve}_\theta-\text{target}_\theta|<0.1)}$\\

\hline \multirow{2}{*}{In-hand Reorientation} & 0.1 & 46 & 24 & $-7||\text{cube}_\text{rpy}-\text{target}_\text{rpy}||$ \\
& & & & $-1000\mathbbm{1}{(\text{isdrop})}$ \\ 

\hline \multirow{3}{*}{Handwriting} & 0.1 & 48 & 24 & $-100||\text{tip}_\text{xy}-\text{target}_\text{xy}||$ \\
& & & & $-20||\text{tip}_\text{z}||$ \\ 
& & & & $-10\mathbbm{1}{(\text{forwardtipping}>0)}$\\

\hline \multirow{2}{*}{Baoding Balls} & 0.1 & 40 & 24 & $-5||\text{objects}_\text{xyz}-\text{targets}_\text{xyz}||$ \\
& & & & $-500\mathbbm{1}{(\text{isdrop})}$ \\ 

\hline
\end{tabular}
\end{table}


\newpage
\section{Tracker Details}
\label{app:tracker}

In order to run this experiment in the real world, we use a camera tracker to produce 3D position estimates for the Baoding balls. The camera tracker serves the purpose of providing low latency, robust, and accurate 3D position estimates of the Baoding balls. To enable this tracking, we employ a dilated CNN modeled after the one in KeypointNet~\citep{Suwajanakorn:2018}. The input to the system is a 280x180 RGB stereo pair (no explicit depth) from a calibrated 12 cm baseline camera rig. The output is a spatial softmax for the 2D location and depth of the center of each sphere in camera frame. Standard pinhole camera equations convert 2D and depth into 3D points in the camera frame, and an additional calibration finally converts it into to the ShadowHand's coordinates system. Training of the model is done in sim, with fine-tuning on real-world data. Our semi-automated process of composing static scenes with the spheres, moving the strero rig, and using VSLAM algorithms to label the images using relative poses of the camera views substantially decreased the amount of hand-labelling that was requiring.  We hand-labeled only about 100 images in 25 videos, generating over 10,000 training images. We observe average tracking errors of ~5 mm and latency of ~20 ms, split evenly between image capture and model inference. 